%% file: IEEE_Trans_main.tex
\documentclass[10pt,journal,compsoc]{IEEEtran}

\usepackage[ruled,linesnumbered]{algorithm2e} 
\usepackage{graphicx} 
\usepackage{subfig} 
\usepackage{amsfonts}
\usepackage{amssymb}
\usepackage{amsmath}
\usepackage{url}
\usepackage{mathrsfs}
\usepackage{multirow}
\usepackage{booktabs} 
\usepackage{makecell}
\usepackage{threeparttable} 
\usepackage{cite}
\usepackage{bm}
\usepackage{color}

\usepackage{algorithmic}
\usepackage{ulem}
\usepackage{ragged2e}
\renewcommand{\justify}{\leftskip=0pt \rightskip=0pt plus 0cm}

\begin{document}

\title{Counter-Empirical Attacking based on Adversarial Reinforcement Learning for Time-Relevant Scoring System}

\author{Xiangguo~Sun,
        Hong Cheng,
        Hang Dong\IEEEauthorrefmark{2},
        Bo Qiao,
        Si Qin,
        Qingwei Lin
\IEEEcompsocitemizethanks{
\IEEEcompsocthanksitem Xiangguo~Sun and Hong Cheng: The Chinese University of Hong Kong, Hong Kong. \{xgsun,hcheng\}@se.cuhk.edu.hk.  
        
\IEEEcompsocthanksitem Hang Dong, Bo Qiao, Si Qin, and Qingwei Lin: Microsoft Research, China. \{hangdong,boqiao,si.qin,qlin\}@microsoft.com

}
\thanks{\IEEEauthorrefmark{2} Hang Dong is the corresponding author.}
}

\markboth{IEEE TRANSACTIONS ON KNOWLEDGE AND DATA ENGINEERING}%
{Xiangguo Sun \MakeLowercase{\textit{et al.}}: }

\IEEEtitleabstractindextext{%
\begin{abstract}
\justify{Scoring systems are commonly seen for platforms in the era of big data. From credit scoring systems in financial services to membership scores in E-commerce shopping platforms, platform managers use such systems to guide users towards the encouraged activity pattern, and manage resources more effectively and more efficiently thereby. 
To establish such scoring systems, several "empirical criteria" are firstly determined, followed by dedicated top-down design for each factor of the score, which usually requires enormous effort to adjust and tune the scoring function in the new application scenario. What's worse, many fresh projects usually have no ground-truth or any experience to evaluate a reasonable scoring system, making the designing even harder.   
To reduce the effort of manual adjustment of the scoring function in every new scoring system, we innovatively study the scoring system from the preset empirical criteria without any ground truth, and propose a novel framework to improve the system from scratch. In this paper, we propose a "counter-empirical attacking" mechanism that can generate "attacking" behavior traces and try to break the empirical rules of the scoring system. Then an adversarial "enhancer" is applied to evaluate the scoring system and find the improvement strategy. By training the adversarial learning problem, a proper scoring function can be learned to be robust to the attacking activity traces that are trying to violate the empirical criteria. Extensive experiments have been conducted on two scoring systems including a shared computing resource platform and a financial credit system. The experimental results have validated the effectiveness of our proposed framework.}

\end{abstract}

}
\maketitle

\input{intro_2}
\input{preliminaries_2}

\input{method_2}

\input{experiment}

\input{related_work}
\input{Conclusion}

\bibliographystyle{IEEEtran}
\bibliography{ref}
 \input{appendix}
 \vspace{-0.5in}
\input{author}
\end{document}

%% file: intro_2.tex
\section{Introduction}

{User scoring is originally used in finance and banking \cite{wang2020using} and has played an important role in financial risk management. With the rise of various online platforms, user scoring systems have indicated more promising contributions to a broader range of applications such as modern enterprise management \cite{teerasoponpong2022decision}, health risk appraisal \cite{israel2014credit}, online social networks \cite{guo2016personal}, etc. }

\begin{figure}[t]
\centering
\subfloat[credit in bank]{
\label{fig:t1q2}
\includegraphics[width=0.24\textwidth]{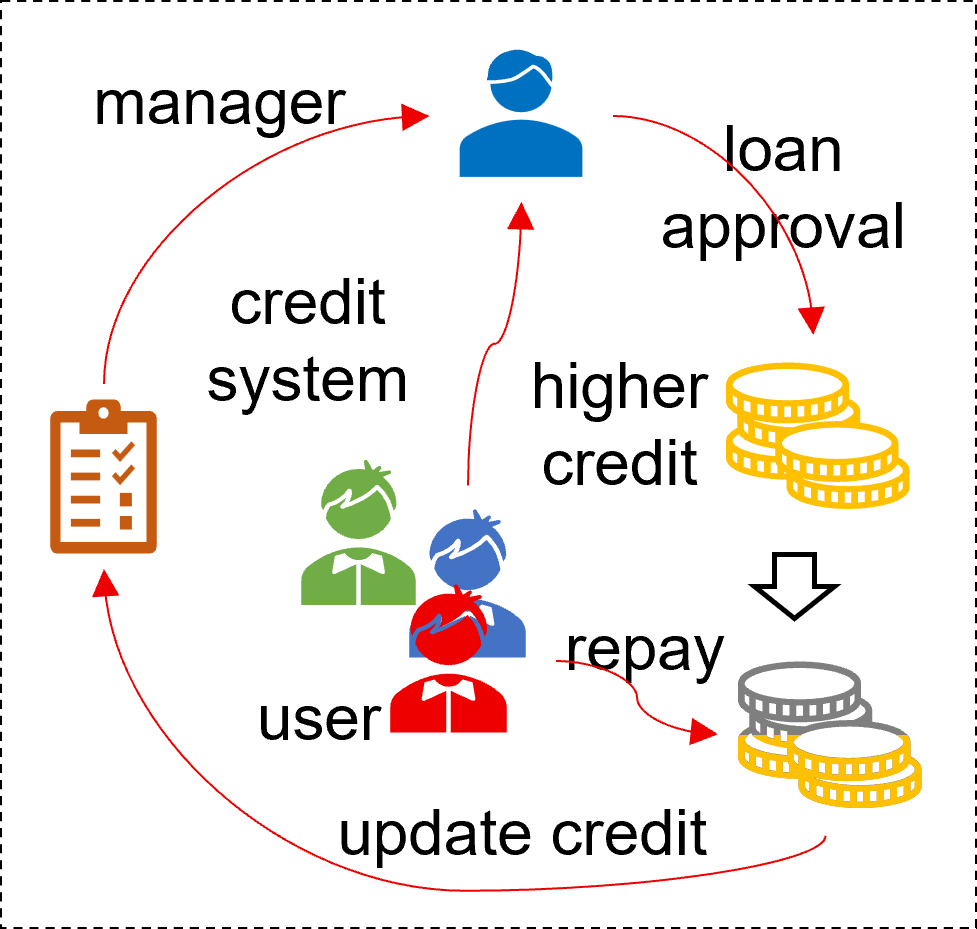}%
} 
\subfloat[credit in online forums]{
\label{fig:t1q1}
\includegraphics[width=0.24\textwidth]{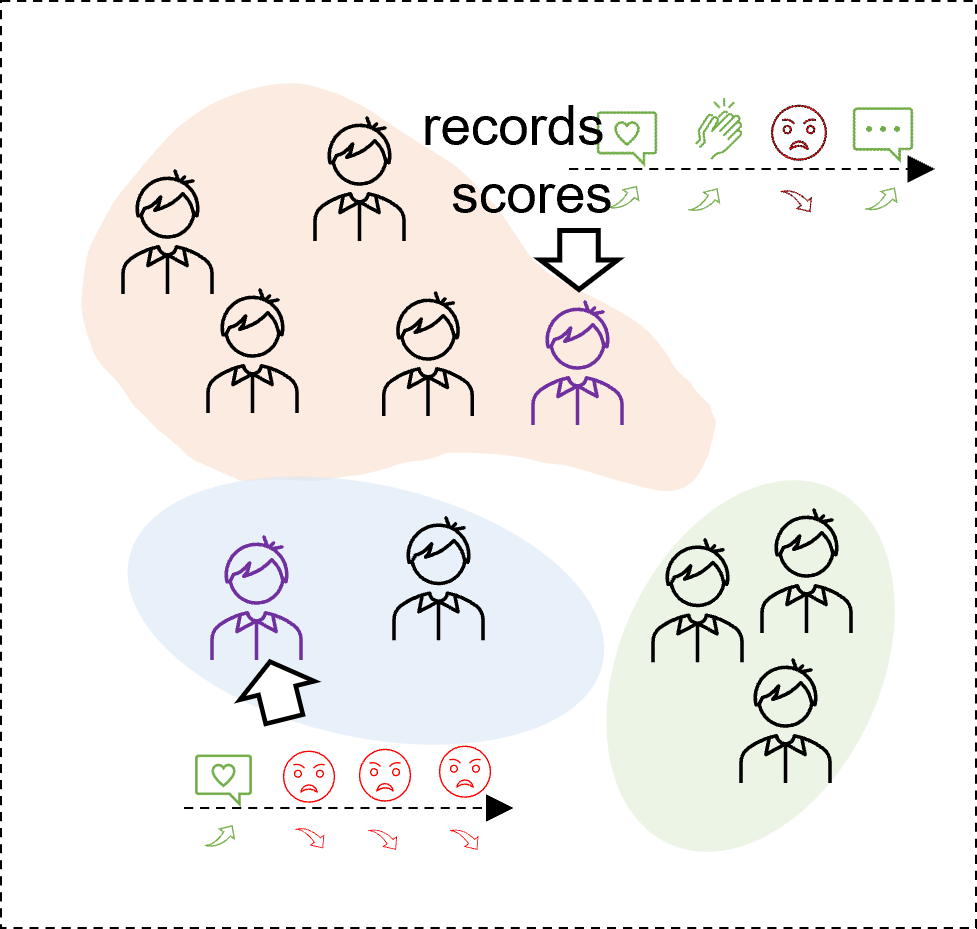}%
}
\caption{Credit Scoring Systems in Bank and Online Forums}\vspace{-3ex}
\label{fig:credit_case}
\end{figure}

In practice, the scoring system usually evaluates users' behavior records and scores them following some "empirical criteria". These empirical criteria are some preferences over the possible activities on the platform for better efficiency and sustainability from the perspective of the platform.
For example, the bank prefers users to repay in time and thus the empirical criterion can be to "raise the credit scores for these users that repay in time" {(as shown in Figure \ref{fig:t1q2})}. Similarly, a scoring system of online forums shown in Figure \ref{fig:t1q1} may encourage the members to be active and polite, and then an empirical criterion can be "to raise a user's score if she/he creates a post". Usually, a user with a higher credit score is more likely to obtain a higher reputation or become easier to access resources. {With the recent rapid development of artificial intelligence techniques, there are plenty of works applying machine learning methods on several supervised tasks related to scoring systems, such as loan rejection prediction\cite{maldonado2010semi} and user classification \cite{li2009hybrid}. However, there are more emerging cases that are entirely fresh due to various demands. Existing work focusing on familiar domains with reliable knowledge is far from sufficient to deal with those entirely new scoring functions without too much experience. }

{
Configuring the scoring function in an unfamiliar application is never easy. \textbf{First}, many user historical activities are commonly used to form time-relevant factors that would influence the user score \cite{thomas2017credit, dumitrescu2022machine}. These time-relevant factors usually form different modules of the final score, which are difficult to be well taken care of using limited experts' experience. \textbf{Second}, in many entirely new business cases, we usually need to evaluate the feasibility of a proposal before actually deploying it. In this case, we may have to start the scoring system from scratch without the benchmark for a scored value. Although we could collect some pseudo labels or feedback from some offline surveys such as the A-B test, they are highly time-consuming and demand huge manpower, and thus it is very hard to cover most user activity traces for an objective evaluation in the business planning stage. Therefore, a more reliable evaluation method relying on fewer supervised signals is urgently needed to solve this challenge. \textbf{Last but not least}, the empirical criteria can be violated by deliberately designed activity traces. For example, in the financial credit scoring system that encourages frequent credit card repayments and higher quota consumption, it is still possible to violate the empirical rule that "having frequent repayments should lead to a higher credit score" by frequent repayments but lower quota consumption when these two factors are added with specific weights}. The effect of past activities on the score can be hard to get properly represented, which can also cause the problem of violation on the empirical criteria designed for the system \cite{kozodoi2022fairness}. 


{In this paper, we wish to go beyond traditional supervised credit scoring approaches and propose the counter-empirical attacking method to help design more }reliable scoring functions that are robust to attacking behaviors on the preset empirical criteria and provide a practical scoring mechanism in new application scenarios without heavy dependence on expert knowledge. {To overcome the challenges mentioned above, we propose a reinforcement-learning-based attacker that tries to find a policy to get higher scores but violate the empirical criteria of the scoring system by taking care of real user behavior patterns and the vulnerability of time-relevant factors. Through this "counter-empirical" attack, our model can generate sufficient simulated behavior traces to find the weakness of the credit scoring system and evaluate its reliability effectively without huge manpower costs. We further design a gradient-based enhancer to remedy the vulnerability of the credit system and better resist various counter-empirical attacks.} Our major contributions in this paper are summarized as follows:

\begin{itemize}
    \item To the best of our knowledge, this is the first work trying to design the scoring systems from scratch without using supervised learning techniques but with deep reinforcement learning techniques.
    \item We propose a counter-empirical attacking method to evaluate a credit scoring system, where the learned attacker aims to both mimic true user activities and generate activities of counter cases that break the preset empirical criteria of the scoring system. 
    \item Based on our proposed attacking model, we further set an enhancer to adjust the parameters in the scoring function to enhance its robustness to attacks. 
    \item Extensive experiments and two real-world applications further demonstrate the effectiveness of our proposed framework, and show that our framework is generally effective across different applications.
\end{itemize}

{The rest of this paper is organized as follows: 
We introduce the preliminaries in section \ref{sec:pre}. Then we present our approach in section \ref{sec:attacker} with extensive experiments in section \ref{sec:exp}. Section \ref{sec:related} reviews related work and section \ref{sec:con} concludes the paper.}

%% file: preliminaries_2.tex
\section{Preliminaries}\label{sec:pre}
In this section, we first present the necessary concepts and then formulate our problem.

{\textbf{Reinforcement Learning:} Reinforcement learning is used to solve the Markov decision process (MDP) denoted by $(\mathcal{S},\mathcal{A},T,R)$ where $\mathcal{S}$ is the state space, $\mathcal{A}=\{a_1,a_2,\cdots,a_n\}$ is the action space, and $T$ is the transition function where $T(s,a,s^{\prime})$ denote the probability of changing to state $s^{\prime}$ given action $a$ in state $s$. $R$ is the reward of a specific state and action pair. The agent is usually represented by a policy like $\pi(a|s)$ which indicates the possible action $a$ when it observes $s \in \mathcal{S}$. Different states usually bring about different expected rewards in the future and the value can be estimated by a value function $V(s)$, which means the expected cumulative discounted returns the agent will have in the future after reaching state $s$. The cumulative discounted returns are defined by: $r_t+\gamma r_{t+1}+\gamma^2 r_{t+2}+\cdots$ where $r_t$ is the reward at time $t$ and $\gamma$ is the discount factor.

}

\textbf{Scoring Function:} A typical scoring function for user $i$ at time step $t$ is the sum of several time-relevant factor modules:
\begin{equation}
\label{eqn:score_form}
    S_i^t = \sum_{k=1}^K w_k \cdot m_{k}(t, \phi_k, \mathcal{H}^t_i)
\end{equation}
where $w_k$ is the weight for the $k^{\textrm{th}}$ factor module and $m_{k}(t, \phi_k, \mathcal{H}^t_i)$ is the module score for the $k^{\textrm{th}}$ time-relevant factor module, with $\phi_k$ being the parameters inside the module function and $\mathcal{H}^t_i = \{a_t, a_{t-1}, \ldots\}$ is the historical activities of user $i$ for calculating each module score. {In Appendix \ref{sec:app}, we present two examples of scoring functions in different applications. Note that the form of the scoring function is not restricted to Equation \eqref{eqn:score_form}. We use this form to illustrate the proposed framework because it is widely used in many related business areas. }

\textbf{Empirical Criteria:} In the designing stage of a scoring function in the form of Equation \eqref{eqn:score_form}, there is a set of empirical criteria in the following form: 
\begin{equation}
\label{eqn:empirical_cri}
    S_i^t < S_i^{t+1} \textrm{ if } a_i^t \prec a_{i}^{t+1} 
\end{equation}
where $a_i^t$ and $a_i^{t+1}$ represent the activities for user $i$  at time $t$ and $t+1$ respectively, and $\prec \subset \mathcal{A}^2$  represents a partial order for a pair of activities. Since these empirical criteria are applied to two activities of a user, they can effectively encode the intuitive expectations to the scoring systems. If Equation (\ref{eqn:empirical_cri}) fails to be satisfied, we call there is a violation of this empirical criterion.

\textbf{Objectives:} With the above concepts, our goal is to learn the parameters  $\Phi = (\bm w, \bm \phi)$, where $\bm w = (w_1,\ldots,w_K)$, and $\bm \phi = (\phi_1,\ldots, \phi_K)$  for the scoring function so as to minimize the expected frequency of counter cases weighted over their severity against the preset empirical criteria. {We formulate this problem as a two-player zero-sum game: we design an attacker with parameter $\theta$, and an enhancer with parameter $\beta$. The attacker interacts with the scoring system $\Phi$ to generate as many counter-empirical cases as possible, then the enhancer adjusts the scoring system to avoid such an attack. Let $v(\beta,\theta)$ be the value of the game, then we wish to obtain the robust version of the scoring system with $\Phi^*$ when the attacker and the enhancer come to Nash equilibrium:
$
    \Phi^{*}=\arg_{\Phi} \min_{\beta} \max_{\theta} v(\beta, \theta;\Phi)
$
}

%% file: method_2.tex
 \section{Adversarial Learning for Scoring}\label{sec:attacker}

\begin{figure}[ht]
    \centering
    \includegraphics[width =0.98\linewidth]{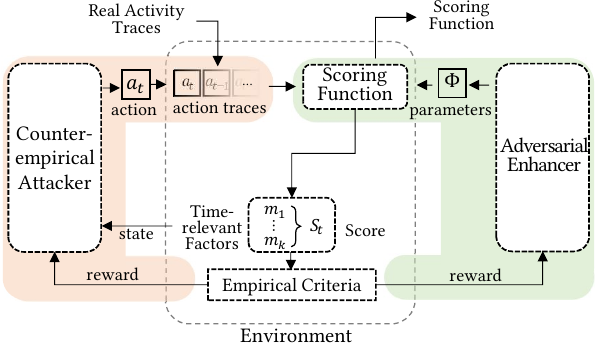}
    \caption{Our proposed adversarial learning framework.}\vspace{-3ex}
    \label{fig:framework}
\end{figure}

The overview of our framework is shown in Figure \ref{fig:framework}. For the counter-empirical attacker, it tries to learn a policy that can generate an activity trace resulting in as many severe counter cases against the preset empirical criteria as possible under the current scoring function. The adversarial enhancer then updates the parameters of the scoring function to prevent the attack. In the following, we first introduce the reinforcement learning framework for the counter-empirical attacker {in section \ref{subsec:frame}}, then explain {how to learn the counter-empirical policy in section \ref{subsec:acnn}}, and last we present how the enhancer works as well as the adversarial learning algorithm designed for getting an approximate optimal solution under this learning problem as a two-player zero-sum discounted game {(section \ref{sec:booster})}.


\subsection{Reinforcement Learning Framework for Attacker}\label{subsec:frame}

In our deep reinforcement learning framework for attacker, the goal is to maximize the expected frequency of counter cases weighted over their severity against the preset empirical criteria in the scoring system. In this framework, the scoring function is fixed, and the attacker is trained with a hybrid actor-critic neural network. For the attacker agent, the environment evaluates the activity traces of the attacker, calculates the score value and then returns all values from those time-relevant factor modules as a state vector. Then the attacker observes the state vector and decides the next action trying to violate the empirical criteria. In our problem setting, users' actions on most platforms usually contain both discrete actions (such as "applying for quota or not") and continuous actions (such as "applying for quota with a specific amount"). Therefore, we choose to use the actor-critic structure due to its effectiveness to deal with both discrete action space and continuous action space \cite{bhatnagar2007incremental,grondman2012survey}. In this actor-critic framework, the actor is used to generate new actions from the agent and the critic is used to calculate the gradient for updating the actor-network.    

{\textbf{Hybrid Actor-Critic Neural Networks}} Although the traditional actor-critic framework \cite{sutton2018reinforcement} can deal with discrete actions or continuous actions, it is not designed for the discrete action followed by a continuous action, which is the target scenario in this paper. A straightforward way is to join several actors independently for different types of actions, and then use a switch to control the channel for each type of action. 
However, this framework suffers from at least twofold imperfections: First, some channels may be not trained well. For example, in a cloud service platform, a user may ask some quota at the beginning and then keep using them in the following steps. The parameters in the "ask" channel have only one time to learn but the gradients in the other channel can accumulate and update their parameters better. Second, the actor nets from different channels usually share similar network structures, which is parameter redundant and learning inefficient.

\begin{figure}[h]
\centering
\includegraphics[width=0.5\textwidth]{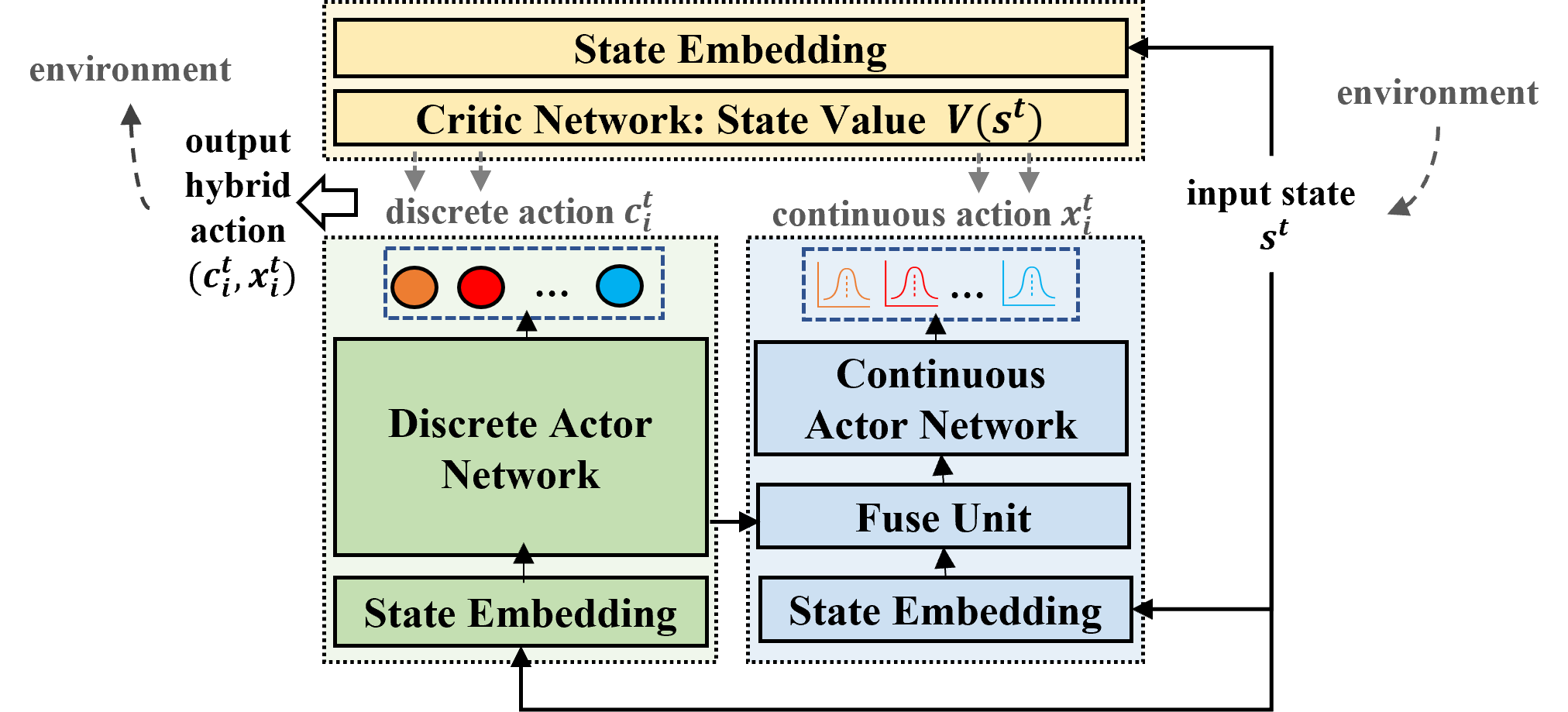}
\caption{{Hybrid Actor-critic Neural Networks.}} \label{fig:actor_critic}
\end{figure}

To this end, we compress these continuous action channels with one uniform actor network and use a separate discrete actor as a prompt to inspire the continuous actor to generate corresponding values for different types of action. In this way, we can reduce the parameter complexity dramatically and learn more effectively in this sparse action space. As shown in Figure \ref{fig:actor_critic}, our actor network is separated into a discrete actor and a continuous actor.  The state vector $s^{t}$ is fed into a state embedding module for both the actor and critic networks. The discrete actor network will output a probability distribution over all possible discrete actions, which will be fed into the continuous actor network and combined with the state embedding. Through this structure, the continuous actor network can learn the specific policy for continuous actions related to each possible discrete action.  The critic network will calculate the advantage according to the final action and the state vector, which can guide the actor to choose better attacking actions. 
The details of the critic network, actor network, and training algorithm will be explained in the following part.

\textbf{State}
The attacker agent aims to violate the empirical criteria, with the calculated scores as feedback from the scoring system, and these scores are the states in our proposed reinforcement learning framework. More specifically, when there are multiple time-relevant factor modules in the scoring function, we stack them into a state vector:
\begin{equation}
    S^t  =  (m_{1}^t, \ldots, m_{N}^t)
\end{equation}
where $m_{k}^t \in \mathbb{R}, k = 1,\ldots, N$ are the values of time-relevant factor modules at time step $t$.

\textbf{Action}
The action of the attacker agent at time $t$ is the next activity on the platform, including which discrete action to take and the following continuous amount of this discrete action, such as first to decide to repay or take out loans, followed by the amount of the repayment or the loan:
\begin{equation}
    a^t = (c^t, x^t) \in \mathcal{A}
\end{equation}
where $c^t$ and $x^t$ represent the discrete action and the following continuous-valued action respectively. 
The action space:
\begin{equation}
    \mathcal{A} = \cup_{c\in \mathcal{A}_c} \{(c, x)| x \in \mathcal{X}_c\}
\end{equation} 
where the discrete action space is $\mathcal{A}_c=\{c_1,\ldots, c_M\}$ with $M$ being the total number of choices for the discrete action, and the continuous action space corresponding to discrete action $c$ is $ \mathcal{X}_c \in \mathbb{R}$. Without loss of generality, we assume each user would only take one type of such discrete actions at each time step. 


\textbf{Reward}
As the attacker aims to generate activity traces that can violate the preset empirical criteria, we design to give a reward to the attacker agent every time when there's a violation case towards any empirical criteria. Assume there is a set of empirical criteria in the form of equation \eqref{eqn:empirical_cri}. For each time step and for each empirical criterion, if there's a violation of this criteria, the attacker will be given a positive reward. A penalty is accordingly calculated if the attacker fails to generate any violation cases. {In the Appendix \ref{sec:app}, we present the detailed formulation of our designed rewarding functions for the two different applications.}




{\subsection{Policy Learning for Counter-empirical Attacking}\label{subsec:acnn}
\subsubsection{Critic Learning}}
The critic network calculates the advantage function based on the current state and the output of the actor network.
Specifically, we use the Generalized Advantage Estimator (GAE) \cite{schulman2015high} to calculate the advantage of the given state and action:
{\begin{equation}\label{eq:gae}
    \hat{A}_t=\delta_t+\gamma \lambda \delta_{t+1}+\cdots+(\gamma \lambda)^{T-t+1}\delta_{T-1}
\end{equation}
where $\gamma$ is the discount factor, and $\delta_t$ is defined by $\delta_t=r_t+\gamma V(s_{t+1})-V(s_{t})$.} Here $V(s)$ is the approximated state value given by the critic network and $r_t$ is the reward obtained at time step $t$.   
The actor in the attacker then reads the states from the scoring system and determines the next action according to this advantage value given by the critic.


To estimate the state value, we design the critic network as a two-layer multi-layer perceptron (MLP) network and the parameters of the critic are denoted as $\theta^{\mathit{critic}}$. To learn these parameters, we  minimize the mean-squared error loss between the estimated values and the discounted return:
\begin{equation}\label{eqn:critic_loss}
    \mathcal{L}_{\mathit{critic}}=\frac{1}{T}\sum_{t=0}^T\left(V(s_t;\theta^{\mathit{critic}})-R_t\right)^2
\end{equation}
where {$R_t=\sum_{k=0}^{T} \gamma^k r_{t+k+1}$} is the discounted return of the attacker.

{\subsubsection{Actor Learning}}

Different from the classic actor-critic framework, the action space in our setting is a hybrid of discrete and continuous action space. Therefore, we design a discrete actor and a continuous actor in our actor network. The discrete actor has an independent state embedding layer to encode the states from the environment. Then we use a two-layer MLP to output the probability distribution on the candidate discrete actions as follows:
\begin{equation}\label{equ:dis_probs}
\mathbb{P}_t(c_1), \ldots, \mathbb{P}_t(c_M) \leftarrow {Actor_{\mathit{dis}}(s_t; \theta^{\mathit{dis}})}    
\end{equation}
where $Actor_{\mathit{dis}}(s_t; \theta^{\mathit{dis}})$ is the output of the discrete actor network parameterized by $\theta^{\mathit{dis}}$ given state $s_t$,  $\mathbb{P}_t(c_m)$ is the probability of choosing action $c_m \in \mathcal{A}_c$ given $s_t$ at the $t^{\textrm{th}}$ step. For all the discrete actions, we have $\sum_{m=1}^M\mathbb{P}_t(c_m)=1$.

The continuous action followed by the discrete action will then be generated via a Gaussian distribution:  $x_m\sim\mathcal{N}(\mu_m,\sigma)$
where $\sigma$ is a hyper-parameter controlling the variance of the Gaussian distribution and the mean value $\mu^t_m$ for the $m^{\textrm{th}}$ discrete actor $c_m$ is obtained from the output of the continuous actor network:
\begin{equation}
   \!\!\!  (\mu^1_t,\ldots,\mu^M_t)\leftarrow {Actor_{\mathit{con}}\Big(s_t,  \mathbb{P}_t(c_1), \ldots,\mathbb{P}_t(c_M) ;\theta^{\mathit{con}}\Big)}
\end{equation}
where $Actor_{\mathit{con}}\Big(s_t,  \mathbb{P}_t(c_1), \ldots,\mathbb{P}_t(c_M);\theta^{\mathit{con}}\Big)$ is the output of the continuous actor network parameterized by $\theta^{\mathit{con}}$ given the state $s_t$ and the output probability distribution of the discrete actor network. 
Different from the discrete actor, the continuous actor takes both the state embedding and the discrete action probability distribution as input because continuous actions not only depend on the given state but also the action type (discrete actions). 

In practice, we can first sample a discrete action $c_m$ according to the distribution over all the possible discrete actions, then the sampled discrete action is treated as a mask to prevent the update on the distribution of other continuous actions with regard to other discrete actions, except the $m^{\textrm{th}}$ distribution $\mathbb{P}_t(c_m)$. Then we can further sample a continuous value $x_m$ for the discrete action $c_m$. The complete action $a_m = (c_m, x_m)$ is thereby obtained.

To learn the policy for the attacker agent effectively and efficiently,
we use the Proximal Policy Optimization (PPO) \cite{schulman2017proximal} to train the reinforcement learning model for the attacker. PPO learns the stochastic policy via minimizing the following function:
\begin{equation}
\label{eq:ppo}
\begin{aligned}
   \!\!\! &\mathcal{L}_{\mathit{PPO}}(\theta)\! =\\
    \!\!\!&\!-\mathbb{E}_t\!\left[ 
    \hat{A}^t \min\left(\!
\frac{\pi_{\theta}(a|s_t)}{\pi_{\theta_{\mathit{old}}}(a|s_t)},
\text{clip}(\frac{\pi_{\theta}(a|s_t)}{\pi_{\theta_{\mathit{old}}}(a|s_t)},\! 1\! -\! \epsilon,\! 1\!+\!\epsilon )
\!\right)
\!\right]
\end{aligned}
\end{equation}

 where $\pi_{\theta_{\mathit{old}}}$ is the action distribution generated by the old policy, $\pi_{\theta}$ is the action distribution estimated by the new policy, and $ \epsilon$ is the hyper-parameter to control the degree of evolving from the old policy to the new policy. {In this paper, both the discrete actor and the continuous actor have a similar objective function defined by equation \eqref{eq:ppo}. 
That means, for the discrete actor,  $\pi^{dis}_{\theta^{dis}}$ is reformulated by $Actor_{dis}$ as in equation \eqref{equ:dis_probs}; For the continuous actor,  $\pi^{con}_{\theta^{con}}$ can be obtained by $Actor_{con}$.} Then we can get the loss function for the continuous actor as $\mathcal{L}_{\mathit{con}}\!=\!\mathcal{L}_{\mathit{PPO}}(\theta^{\mathit{con}})$, and $\mathcal{L}_{\mathit{dis}}\!=\!\mathcal{L}_{\mathit{PPO}}(\theta^{\mathit{dis}})$ for the discrete actor.

\subsubsection{Regulating the Attacker via Real Traces}\label{subsec:simulation}
Apart from trying to violate the empirical criteria, the attacker is also supposed to generate the activity traces that are more likely to really happen for real users, because the scoring system is finally for real users and we don't need to search for the policy space that's distant to the real policy set. 
Collection of such real activity traces is common nowadays for most applications and can be used to regulate the attacker to generate more realistic activity traces.
An example of such real activity traces is the records of repayment and loan taking for customers from banks.  To better utilize these activity traces in training our attacker agent,  we propose a regulation component in the training network. The main idea of this regulation component is to reduce the distance between the attacker's activity trajectories and the real users' activity records. 

To deal with the temporal sequences of the generated trajectories and users' activity records with variable length, we introduce the {soft dynamic time warping (a.k.a Soft-DTW)} \cite{cuturi2017soft} as the distance metric to align two sequences with different lengths via warping path. The basic idea of dynamic time warping (DTW) is to solve the minimal-cost alignment problem between two sequences using dynamic programming. Soft-DTW is a smoothed formulation of DTW and is a differentiable loss function, whose value and gradient can be computed efficiently.
Specifically, let $\mathbf{p}=(p_1,\cdots,p_n) \in \mathbb{R}^{d\times n}$ and $\mathbf{q}=(q_1,\cdots,q_m)\in \mathbb{R}^{d\times m}$ be two time series with length $n$ and $m$, respectively. We can treat $(p_i,q_j)$ as a point and use a binary matrix $\mathbf{H}\in \{0,1\}^{n\times m}$ to denote one alignment path from $(p_1,q_1)$ to $(p_n,q_m)$ where $H_{i,j}=1$ if $(p_i,q_j)$ is on the path and $H_{i,j}=0$ vice versa. Let $\mathcal{H}_{n,m}$ be the set of these binary alignment matrices where each alignment matrix $\mathbf{H} \in \mathcal{H}_{n,m}$ contains one candidate path from $(p_1,q_1)$ to $(p_n,q_m)$. Let $\mathbf{\Delta}(\mathbf{p},\mathbf{q})$ be the cost matrix where each element is denoted as $\delta (p_i,q_j): \mathbb{R}^d\times \mathbb{R}^d\rightarrow \mathbb{R}_{+}$. Then Soft-DTW between $\mathbf{p}$ and $\mathbf{q}$ is defined as follows:
\begin{equation}
  \begin{aligned}
    {DTW}_\gamma(\mathbf{p},\mathbf{q})=&
         min^\gamma\{<\mathbf{H},\mathbf{\Delta}(\mathbf{p},\mathbf{q})>,\mathbf{H} \in \mathcal{H}_{n,m} \}\\
         =&-\gamma\log\left(\sum\limits_{\mathbf{H} \in \mathcal{H}_{n,m}}e^{-\frac{<\mathbf{H},\mathbf{\Delta}(\mathbf{p},\mathbf{q})>}{\gamma}}\right)
    \end{aligned}
\end{equation}
where $\gamma>0$ is a smoothing parameter. In this paper, the behavior sequences are a series of hybrid actions containing both a discrete action and a continuous value like $(c_i,x_i)$. To calculate $\delta (p_i,q_j)$, we denote the action tuple  $(c_i,x_i)$ as the one-hot vector $\mathbf{v}\in \mathbb{R}^{1\times M}$ where each element corresponds to the continuous action value. Then $\delta(p_i,q_j)$ is defined as the quadratic Euclidean distance between $p_i$ and $q_j$. Let $\mathbf{p}$ stand for a trajectory from the attacker $\theta$. Let $\mathbf{q}$ be one trace from the real data. Then we can calculate the gradient of $\theta$ as follows:
\begin{equation}
\begin{aligned}
    \nabla_\theta {DTW}_\gamma(\mathbf{p},\mathbf{q})=&\nabla_\mathbf{p} {DTW}_\gamma(\mathbf{p},\mathbf{q})\cdot \nabla_\theta \mathbf{p}\\
    =&\left(\frac{\partial \mathbf{\Delta}(\mathbf{p},\mathbf{q})}{\partial \mathbf{p}}\right)^T \mathbb{E}_\gamma [H]\cdot \nabla_\theta \mathbf{p}
\end{aligned}
\end{equation}
where $\mathbb{E}_\gamma [H]$ is an average of all alignment matrices on  $\mathcal{H}_{n,m}$ via Gibbs distribution $p_\gamma \propto e^{-<\mathbf{H},\mathbf{\Delta}(\mathbf{p},\mathbf{q})>/\gamma}$, which can be efficiently calculated by \cite{cuturi2017soft}.

With the above discussion, we can define the following Soft-DTW loss on all attacking trajectories and the real traces:
\begin{equation}\label{equ:dtw}
    \mathcal{L}_{\mathit{DTW}}=\sum_{x \in \mathcal{B}(\theta)}\sum_{y \in \mathcal{R}}\mathit{DTW}_\eta(x,y)
\end{equation}
where $\eta$ is a hyper-parameter for Soft-DTW controlling the smoothness factor, 
$\mathcal{B}(\theta)$ is the set of trajectories generated by the attacker with parameter $\theta$, and $\mathcal{R}$ is the set of the real activity records. 

\begin{algorithm}[t]
\caption{Adversarial Framework for the Scoring System 
}
\label{alg:adver_alg}
\begin{algorithmic}[1]
\REQUIRE
Attacker; Enhancer; Environment; Real activity traces;
\ENSURE
The final scoring function parameters $\Phi$
\STATE Initialize scoring function parameters $\Phi$;
\STATE Initialize attacker parameter $\theta=\{\theta_0^{\mathit{con}},\theta_0^{\mathit{dis}},\theta_0^{\mathit{critic}}\}$;
\REPEAT

\REPEAT
\STATE Sample attacker trajectories $\mathcal{B}(\!\theta|\Phi)$ with $\theta$  given $\Phi$;
\STATE Calculate $\mathcal{L}_{\mathit{critic}}$, $\mathcal{L}_{\mathit{con}}$, $\mathcal{L}_{\mathit{dis}}$, and  $\mathcal{L}_{\mathit{DTW}}$;
{\STATE Update $\theta^{critic}$ by $\mathcal{L}_{\mathit{critic}}$;
\STATE Update $\theta^{dis}$ and $\theta^{con}$ by $\mathcal{L}_{\mathit{dis}}+\mathcal{L}_{\mathit{con}}+\mathcal{L}_{\mathit{DTW}}$;}

\UNTIL{convergence}
\STATE Generate trajectories $\mathcal{T}$ by the attacker $\theta_{i}$ given $\Phi$.
\REPEAT
{\STATE Calculate the cumulative rewards $R_{\mathcal{T}}(\Phi)$ on $\mathcal{T}$ with $\Phi=({\bm\phi}, \bm{w})$.
\STATE /*The enhancer uses the gradient descent to update scoring function parameters, and minimize the attacking rewards, as opposed to the attacker*/ \\
Treat $\Phi$ as tunable parameter and update it by\\ $\Phi \leftarrow \Phi-r \frac{\partial R_{\mathcal{T}}(\Phi)}{\partial \Phi}$}
\UNTIL{convergence}
\UNTIL{convergence}
\end{algorithmic}
\end{algorithm}

\subsection{Improving Policy for the Scoring System via the Adversarial Enhancer}\label{sec:booster}
 
To enhance the capability of the scoring function against the attacks from the attacker, we adopt an enhancer to adjust the parameters of the scoring function so that the scores can be more robust against the counter-empirical attacker. 
The goal of the enhancer here is just opposite to that of the attacker, and therefore the attacker and the enhancer in this setting can be treated as two players in 
a two-player zero-sum discounted game \cite{Littman1994game}. In this game, the attacker takes actions to get a higher reward, and the enhancer adjusts the parameters in the scoring function to prevent the attacker from getting a positive reward.  {To simplify the policy of the enhancer, we use a fixed policy of optimizing the negative reward of the attacker.} Then we denote the value of the game as $v(\Phi, \theta)$, where $\Phi$ and $\theta$ are the parameters of the scoring function and attacker respectively, then the optimal value $v^*$ is
\begin{equation}
    v^{*}=\min_{\Phi} \max_{\theta} v(\Phi |  \theta)
\end{equation}

%
%
%
%
%
%

\noindent According to \cite{perolat2015approximate}, the above minimax solution exists and it is equivalent to the Nash equilibrium via mixed stationary strategies. Nevertheless, the Nash equilibrium for this game is of exponential complexity in the cardinality of the action spaces. To deal with this intractability, 
we seek to learn stationary policies $\Phi^*$  and $\theta^*$ such that $v(\Phi^*, \theta^*)   \rightarrow v^*$. 

More specifically, we design an iterative algorithm to optimize the policy of attacker and enhancer alternatively. As shown in Algorithm \ref{alg:adver_alg}, the expected output of this algorithm is $\Phi = ({\bm\phi}, \bm{w})$ for all the factor modules in \eqref{eqn:score_form}. In this algorithm, the attacker with parameter $\theta$ will first generate a set of activity trajectories $\mathcal{B}(\theta|\Phi)$ with scoring parameter $\Phi$. Then by updating the actor-critic network introduced in \ref{subsec:acnn}, the attacker agent is updated. After the attacker converges to a stationary attacking policy, this policy is further used to generate a corpus of testing trajectories $\mathcal{T}$ under the current $\Phi$. Following the attacker with parameter $\theta_i$, the enhancer re-calculates the cumulative rewards on $\mathcal{T}$ with variable $\Phi$, which can be denoted as $R_{\mathcal{T}}(\Phi|\theta_i)$ and then the enhancer updates to the optimal $\Phi$ by gradient descent as the updated version of $\Phi$. By iteratively updating the attacker and the enhancer, the reward value stabilizes as the process iterates. The final $\Phi$ after convergence is then output as the learned scoring function parameters.

%% file: experiment.tex
\section{Experiment}\label{sec:exp}

\subsection{Environmental Settings}\label{subsec:exp_set}

As previously discussed, we take two scenarios in the experiment. The first one is the scoring system of an internal cloud computing platform shared by several teams. 
Besides, we also evaluate a concise Bank Credit Scoring system from a public dataset.
For these two applications, we develop the environment separately 
based on OpenAI Gym\footnote{\url{https://gym.openai.com/}}.

{\textbf{Scoring for Cloud Platform:} This application is a cloud computing platform with a fixed amount of capacity shared by the teams as users.
The users can ask for a computing quota in the number of virtual CPU cores and deploy their computing workloads under the limit of the quota approved by the platform administrator. To increase resource utilization, the computing platform encourages the users to use up the approved quota before further asking for more quota. Therefore, the scoring system for this platform scores each user and tends to approve the quota requests from the users with higher scores. There are three types of actions in the scenario: asking for more computing quota (\emph{ask}), deploying under the current quota (\emph{deploy}), and doing nothing (\emph{wait}). See in Appendix \ref{app:azure} for more information.


\textbf{Simulated Financial Credit Scoring:} This is a simplified version of the financial credit scoring system in the real world. 
For simplicity, we consider 4 types of actions: (1) repaying only (\emph{repay}): the user only repays some of his/her debt and does no purchase; (2) consuming only (\emph{consume}): the user makes purchases but does not repay the remaining debt; (3) in-out (\emph{in-out}): the user both repays some debt and does purchase this month; (4) inactivity (\emph{inactive}): the user neither repays nor makes purchases. We consider a nutshell scoring function as shown in Appendix \ref{app:bank}.



\textbf{Hyper-parameters Settings}
We set the clip threshold in Equation \eqref{eq:ppo} as $\epsilon=0.2$; actor learning rate $r_a=0.0001$, critic learning rate $r_c=0.001$, enhancer learning rate $r_e=0.001$, the max length of an episode is set as $100$. In Equation \eqref{eq:gae}, the discount factor $\gamma=0.98$, $\lambda=0.95$. $\sigma$ in the Gaussian distribution of continuous actions is initialized as $0.6$ and reduced by $0.1$ every 10 epochs until it comes to the min value $0.1$. We use the multi-layer perception (MLP) to construct the actor network and the critic network and the network layer number is set as 3. More detailed configurations can be found in our open source code\footnote{\url{https://github.com/sheldonresearch/Microsoft-Scoring-System}}. 
}

\subsection{Evaluation Paradigm}
How to evaluate a scoring system since there is no ground truth? This open problem is a new challenge in our target situation and never happened in the traditional supervised scenario. For example, in a traditional bank credit system, even if it is designed for a new business, we can still borrow some "labels" on users' credits from other banks or similar businesses and then train the model in a supervised method. However, in more wide areas, many projects are entirely fresh and we can refer to very limited annotated datasets. Usually, the credit scoring system is built on some empirical criteria so that the score can encourage preferred behaviors and discourage those opposite behaviors. However, in these new applications, we do not know whether these criteria are properly reflected in the draft credit scoring system without actual deployment.

We have noticed that the key problem preventing a credit scoring system in an unfamiliar area is how to evaluate the system without any ground truth and improve it to be more robust with regard to the designed criteria. This is a common challenge in the practice of establishing a scoring system in many new application scenarios but is currently not well-solved. In light of this, we extensively evaluate our framework by answering the following research questions (RQ): 
{
\begin{itemize}
    \item RQ1: How well does the adversarial training conduct?  
    \item RQ2: How well does the attacker perform? 
    \item RQ3: How well does the enhancer improve the scoring system?
    \item RQ4: How well does the learned scoring system perform? 
\end{itemize}
}

\begin{figure}[htbp]
\centering
\includegraphics[width=0.48\textwidth]{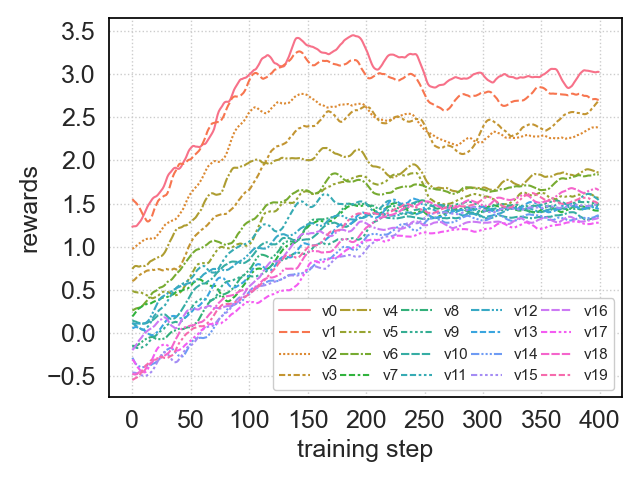}
\caption{Mean Episode Rewards for Attacker.} \label{fig:reward_train_attacker}
\end{figure}

{\subsection{Analysis on the Adversarial Training}
\subsubsection{Attacking Performance in the Adversarial Training}}
For each round of the adversarial iteration, we update the attacker's policy until convergence and each time we record the average training rewards over 10 generated trajectories. Figure \ref{fig:reward_train_attacker} presents the average reward in 20 rounds of adversarial attacking w.r.t the training step. In the plot, each curve denotes the performance of one round of attackers denoted by "v0", ..., "v19", from which we can find that  the attacking reward tends to be stable as the training step continues. Higher rewards mean the attacker can find more violation cases over the preset empirical criteria when attacking the corresponding version of the scoring system. With the scoring system improved it is more difficult for the attacker to find violation cases. This suggests that the learning of the scoring system is effective and the improved scoring system becomes more robust.


\begin{figure}[htbp]
\centering
\includegraphics[width=0.48\textwidth]{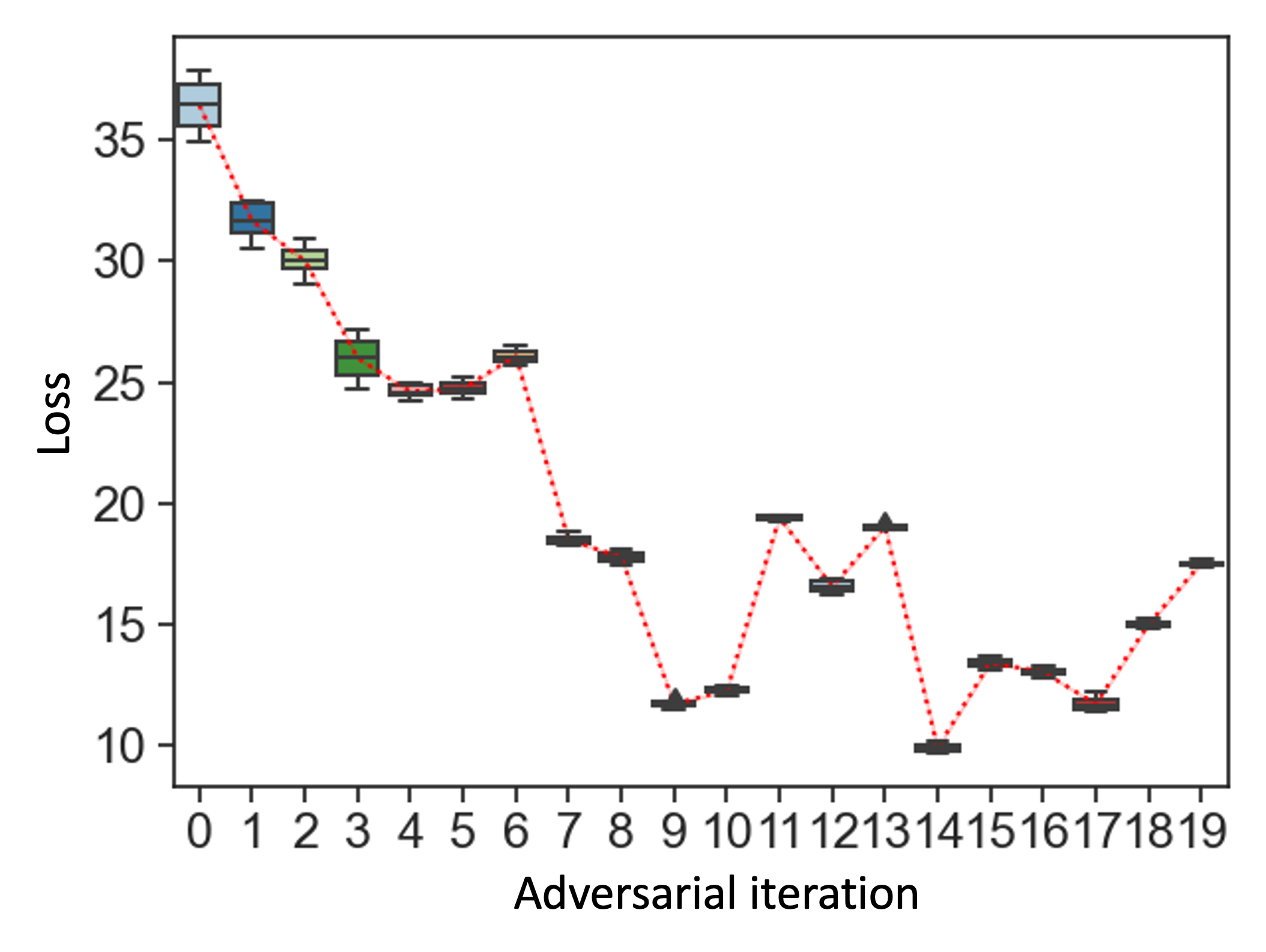}
\caption{{Enhancer Training Loss Curves. Each box contains the training step losses of the enhancer within the same adversarial iteration. The trend of the average loss of each box is denoted by the red dashed line in 20 adversarial iterations along the x-axis.}}
\label{fig:enhancer_loss}
\end{figure}




\subsubsection{Enhancer Convergence Analysis.} 
{As mentioned in section \ref{sec:booster}, the enhancer training loss is just the cumulative rewards on the test trajectories generated by the attacker. The enhancer aims to adjust the parameter of the scoring function to minimize the cumulative rewards, thus making the scoring system more sensitive to the pranking behavior records.
In Figure \ref{fig:enhancer_loss}, we group the enhancer training step losses within each adversarial iteration as a box and present 20 adversarial iterations along the horizontal axis. We can see that the average loss value in each adversarial iteration gradually goes down to a relatively lower level, which means that the scoring system has become more powerful against counter-empirical attacks. The oscillation of this curve indicates the game process of the attacker and the enhancer, which gradually comes to a balanced state. This observation can be also found in the box variances where the early iterations are larger than those in the later iterations. 
}



\begin{figure}[htbp]
\centering
\includegraphics[width=0.48\textwidth]{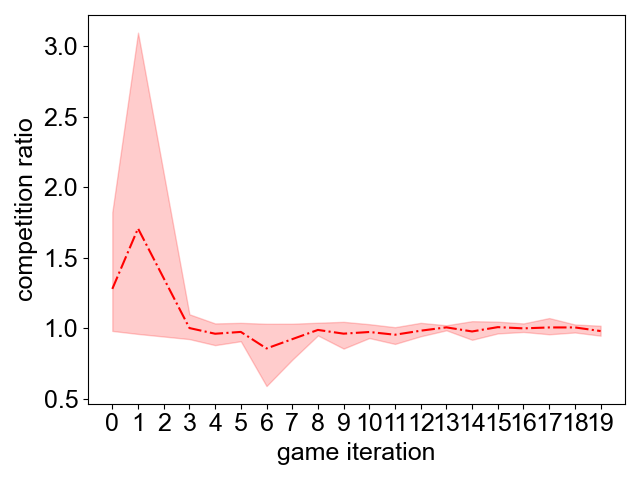}
\caption{Competition Ratio.} \label{fig:competition_ratio}
\end{figure}

\subsubsection{Competition Ratio during Adversarial Learning} 
To evaluate the competition performance of the attacker and the enhancer, 
we first calculate the average attacking rewards $r^{a}_{i}$ on the scoring function in $i^{\textrm{th}}$ round of adversarial training before updated (\textit{step 1}). After the scoring function is updated at the end of $i^{\textrm{th}}$ round of adversarial training (\textit{step 2}), we use the same attacker to calculate the new average attacking rewards $r^{b}_{i}$ on the new scoring system (\textit{step 3}). Then we can define a competition ratio in the $i^{\textrm{th}}$ round as $r^c_i=r^{a}_{i}/r^{b}_{i}$. Intuitively, $r^c_i>1$ means the enhancer beats the attacker in $i^{\textrm{th}}$ round confrontation because the new scoring function reduces the attacker's rewards. On the contrary, $r^c_i<1$ means the attacker beats the enhancer because the attacker achieves higher rewards on the new scoring function. To make the evaluation objective, \textit{step 1}, \textit{step 2}, and \textit{step 3} are all conducted on different trajectory corpora. We further repeat this evaluation 3 times and present the mean values and the variances in Figure \ref{fig:competition_ratio}, from which we can find that the competition ratio gradually comes to $1$ with the adversarial iterations, reflecting the effective convergence of our proposed adversarial learning framework. 

\begin{figure}[htbp]
\centering
\includegraphics[width=0.48\textwidth]{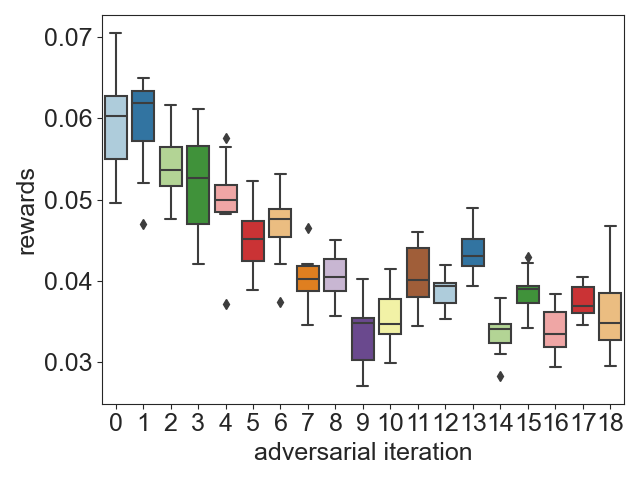}
\caption{Box-plot of Testing Rewards for Different Scoring Systems. } \label{fig:boxplot_test}
\end{figure}

\begin{table}[htbp]
\centering
\caption{Discrete Action Distribution.}
\label{tab:sta_behavior}
\begin{threeparttable}
\renewcommand{\arraystretch}{0.99}
\begin{tabular}{@{}c@{}c@{\hspace{12pt}}r@{\hspace{10pt}}r@{\hspace{10pt}}r@{\hspace{10pt}}r@{\hspace{10pt}}r@{}}
\toprule
   &  & \makecell{Real\\Data} & \makecell{Fully\\Attack}& \makecell{Solely\\Simulate} &\makecell{Solely\\Attack} & \makecell{Random\\Policy} \\ \midrule
\multirow{4}{*}{Cloud} & ask                & 1.48\%      & 9.73\%    & 12.73\%    & 32.73\%    & 25.61\% \\
                       & deploy             & 83.61\%     & 88.64\%   & 76.64\%    & 61.64\%    & 31.59\% \\
                       & wait               & 14.91\%     & 1.63\%    & 10.63\%    & 5.63\%     & 42.81\% \\ 
                       & RMSE\tnote{*}      & 0           & 0.0948    & 0.0803     & 0.2270     & 0.3682  \\
\midrule
\multirow{5}{*}{Bank}  & repay              & 0.14\%      & 12.23\%   & 9.44\%     & 13.02\%    & 23.95\% \\
                       & consume            & 4.16\%      & 0.76\%    & 10.55\%    & 1.39\%     & 25.80\% \\
                       & in-out             & 51.70\%     & 65.37\%   & 59.52\%    & 73.93\%    & 26.30\% \\
                       & inactive           & 44.00\%     & 21.64\%   & 20.49\%    & 11.66\%    & 23.95\% \\
                       & RMSE\tnote{*}      & 0           & 0.1453    & 0.1361     & 0.2070     & 0.2281  \\
\bottomrule
\end{tabular}%
\begin{tablenotes}
\footnotesize
\item[*] RMSE: root mean square error between the variant and the real data.
\end{tablenotes}
\end{threeparttable}
\end{table}

\begin{table}[htbp]
\centering
\caption{Statistics on Continuous Actions.}
\label{tab:sta_con}
\begin{threeparttable}

\renewcommand{\arraystretch}{0.99}
\begin{tabular}{@{}c@{}c@{}r@{\hspace{10pt}}r@{\hspace{10pt}}r@{\hspace{10pt}}r@{\hspace{10pt}}r@{}}
\toprule
   &  & \makecell{Real\\Data} & \makecell{Fully\\Attack}& \makecell{Solely\\Simulate} &\makecell{Solely\\Attack} & \makecell{Random\\Policy} \\ \midrule
\multirow{4}{*}{\makecell{Cloud\\\small{(quota)}}} 
    & \multirow{2}{*}{ask}      &  11.81 &  14.63 &   9.78 &  15.73 &  21.08 \\
    &                           & (5.30) & (3.03) & (3.89) & (5.02) & (5.99) \\
    \cmidrule(l{4pt}){2-7}
    & \multirow{2}{*}{deploy}   &  20.40 &  22.06 &  19.43 &  25.64 &  21.42 \\
    &                           & (8.03) & (5.24) & (4.37) & (7.37) & (5.95) \\ \midrule
\multirow{4}{*}{\makecell{Bank\\\small{(amount)}}}  
    & \multirow{2}{*}{repay}    &  20.02 &  12.66 &  15.73 &  11.37 &  10.63 \\
    &                           & (1.33) & (6.37) & (6.24) & (5.73) & (6.01) \\ 
    \cmidrule(l{4pt}){2-7}
    & \multirow{2}{*}{consume}  &  19.45 &  17.62 &  21.26 &  15.62 &  11.54 \\
    &                           & (1.83) & (7.37) & (3.26) & (9.37) & (8.27) \\
\bottomrule
\end{tabular}

\begin{tablenotes}
\footnotesize
\item Outside the parentheses are mean values, inside are standard deviations. For privacy concerns, we transform the real values in a linear way.
\end{tablenotes}
\end{threeparttable}
\end{table}

\subsection{Attacking Policy Analysis}\label{subsec:aba}

In this experiment, we study how the attacker behaves via analysis on the following policies:

\begin{itemize}
    \item \textbf{Fully Attack.} This is the complete attacking policy proposed in our paper with both the attacking component and the behavior simulation component.
    \item \textbf{Solely Simulate.} This policy  only follows the behavior simulator, which means the attacker does not aims at finding counter cases of empirical criteria, only aims at simulating the activities similar to real users. 
    \item \textbf{Solely Attack}: This policy only aims at finding counter cases of empirical criteria and does not care about whether they are likely to happen in the real world, which means the attacker does not have the simulation component. 
    \item \textbf{Random Policy}: This is a random-policy-based agent that only generates each discrete action with equal probability and continuous actions via the uniform distribution within the preset range according to the real applications.
\end{itemize}


\begin{figure}[htbp]
\centering
\includegraphics[width=0.49\textwidth]{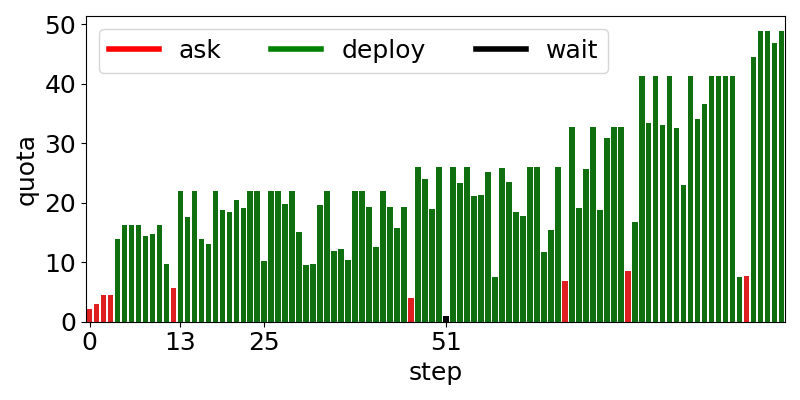}
\caption{An Example of Learned Attacking Trajectory. } \label{fig:attacker_trace}
\end{figure}

We compare the action distribution generated by the above policies and the real datasets\footnote{\url{https://github.com/sheldonresearch/Microsoft-Scoring-System}}, and present the results in Table \ref{tab:sta_behavior}. {Specifically, we generate the same number of trajectories as the real datasets, then we calculate the proportions of different action types (discrete actions).} We also calculated the root mean square error (RMSE) of the discrete action distribution between the variant and the real data to see which policy is most close to the real cases, from which we can see that "Solely Simulate" achieves the closest performance to the real user behavior and "Solely Attack" falls behind. As an integration of "Solely simulate" and "Solely Attack", our "Fully Attack" achieves a balanced simulation between the above two policies. Similar phenomena can be also observed in Table \ref{tab:sta_con}, which presents continuous actions mean values and the standard deviations. 

To further illustrate the attacking policy, we randomly select an attacking trajectory and present it in Figure \ref{fig:attacker_trace}.
Instead of asking large amount of quota all at once, the attacker begins with asking for a small amount of quota for multiple times and tries to avoid the potential penalty. Then it tries to deploy as much as possible to avoid potential penalties on wasting quota. One interesting observation is that the attacker may suddenly deploy at a high level right after a long series of lower-level deployment (see step 13), or suddenly deploy with a few resources just after a long time of higher-level deployments (such as step 25). 
This is a very smart strategy because the attacker raises tougher challenge to the scoring system by suddenly performing a "mischievous" action after a consecutive accumulation of relatively good historical records (or bad record vice versa), which can force the scoring system to be more sensitive and in the meanwhile keep objective to deal with these actions. 

\subsection{Analysis on the Scoring System}\label{subsec:rq3}

\subsubsection{Test Rewards on the Scoring System} We use each round of the attacker to test each round of the scoring system for 10 times and draw the box-plot in Figure \ref{fig:boxplot_test}, from which we can find that the general rewards of the attacker are suppressed with the adversarial iteration continues. That means the trained scoring system is more reliable in preventing bad behaviors.

\begin{figure}[ht]
\centering
\includegraphics[width=0.48\textwidth]{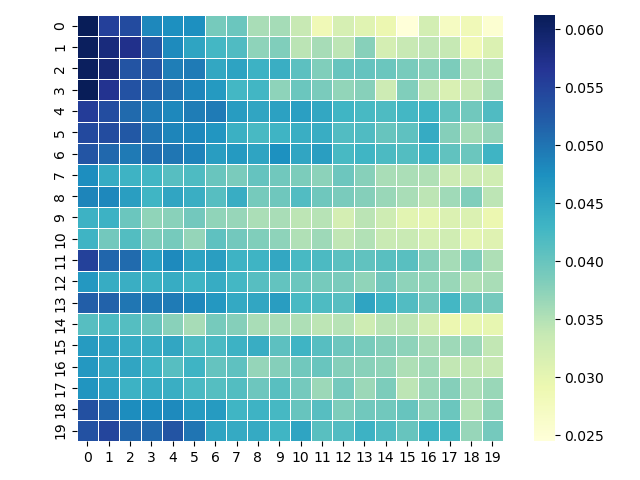}
\caption{Mean attacking rewards across all rounds of attacker and enhancer during the adversarial learning.} \label{fig:heatmap}
\end{figure}

\subsubsection{Cross Evaluation on the Scoring System} 
Here, we use different rounds of attackers to test different rounds of scoring systems and draw the heat map in Figure \ref{fig:heatmap}, where the horizontal axis represent the different versions of the scoring system and the vertical axis represents the different versions of the attacker in the adversarial learning process. Darker blocks stand for higher attacker rewards, from which we can find that later versions of the scoring systems are more robust and reliable against all versions of attackers (lighter color on the right part in Figure \ref{fig:heatmap}).



{\subsection{Ablation Study}\label{subsec:abl}

To verify the effectiveness of our hybrid action space, we replace the hybrid actor with a more straightforward way mentioned in section \ref{subsec:frame}, which uses a switch to control multiple channels for different types of actions (``w/o hybrid''). To verify the effectiveness of the real trace regulating, we remove the Soft-DTW part  (``w/o reg''). We use these variants to conduct the adversarial process and then use the stable scoring function to discriminate between the good users and the bad users. In particular, we take two real-world datasets mentioned in section \ref{subsec:aba}, sample 50 highly good behavior traces and 50 very bad, which are annotated manually. Then we mixed them to see the scores ranked by different scoring systems. From the results presented in Figure \ref{fig:abl}, we can see that our complete framework performs the best while the rest variants can not beat the full model, which indicates the effectiveness of our framework. Moreover, ``w/o reg'' performs the worst, the reason for which might be that its attacking policy is far away from the real user's behavior patterns. 

\begin{figure}[htbp]
\centering
\includegraphics[width=0.48\textwidth]{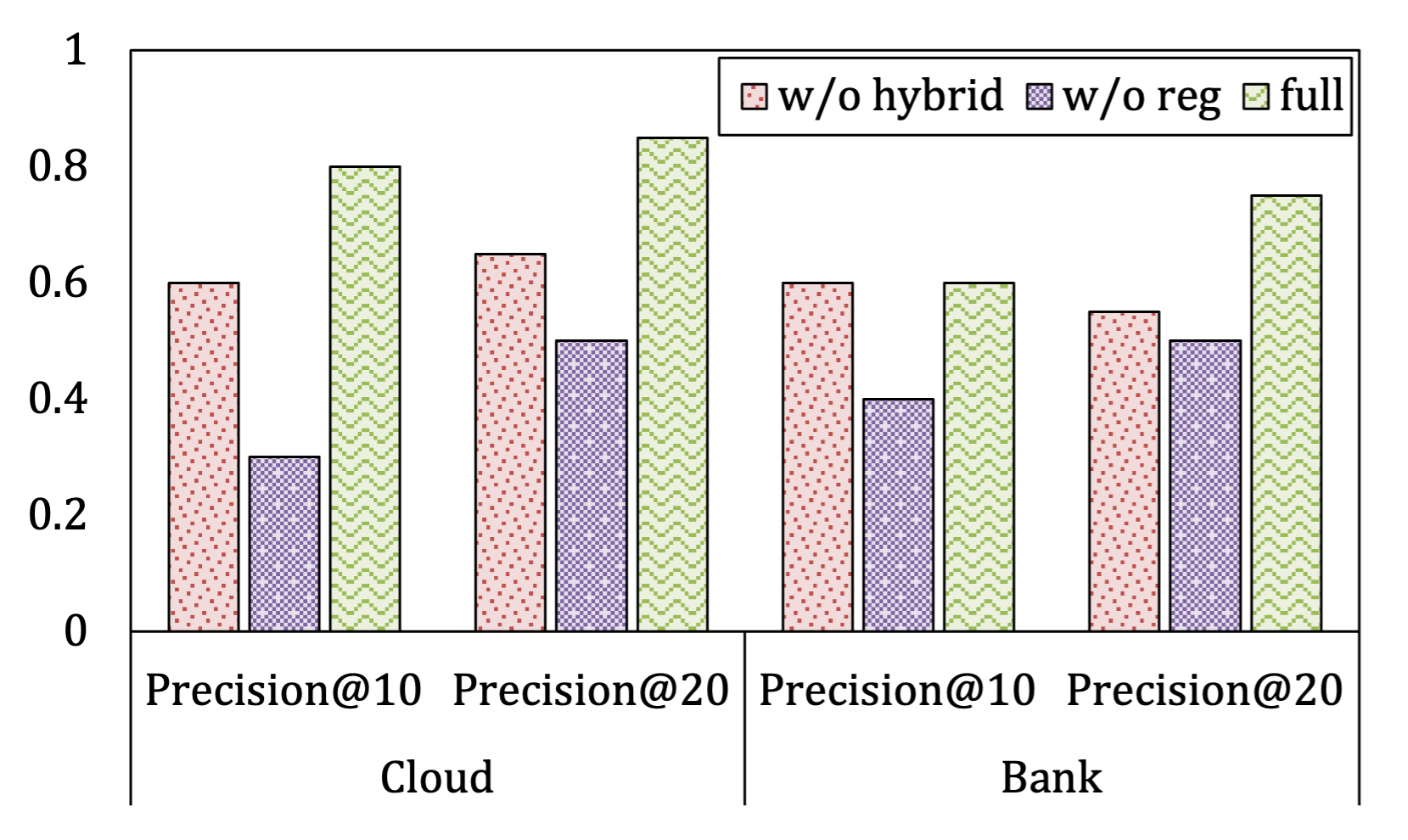}
\caption{Scoring Performance with Different Components.} \label{fig:abl}
\end{figure}

}

\subsection{Scoring Quality Compared with Other Baselines}
Following the setting as section \ref{subsec:abl}, we further evaluate the scoring quality compared with the following baselines: (1) P-DQN \cite{xiong2018parametrized}, which uses Q-learning to generate a distribution on discrete actions and continuous actions. (2) UniWeight, which builds the scoring system with uniform weights. (3) SeqModel, which uses a sequential model like LSTM to learn a latent vector from users' historical records. As shown in Table \ref{tab:baselines}, when we use P-DQN to replace our attacker in the adversarial framework, the learned scoring system can still discriminate between good behaviors and bad behaviors very well, which further demonstrates the superiorities of our counter-empirical adversarial learning. However, P-DQN updates the continuous parameters for all the discrete actions even if they are not included in current choices, which might decrease the performance. In contrast, we separate the discrete action and its continuous values as two actors and design a more effective updating way for these two components, making the performance improve further. The SeqModel and the UniWeight are not good as our method because SeqModel needs sufficient supervised signals but in our cases, the labeled traces are very limited, and the uniform weight of the scoring system is neither good because this scoring function is not designed to avoid deliberate attack.

\begin{table}[htbp]
\caption{Scoring Performance with Different Methods}
\label{tab:baselines}
\resizebox{0.9\columnwidth}{!}{%
\begin{tabular}{@{}ll|lll@{}}
\toprule
                       &           & \makecell[l]{Precision\\@5} & \makecell[l]{Precision\\@10} & \makecell[l]{Precision\\@20} \\ \midrule
\multirow{4}{*}{Cloud} & Ours      & 1           & 0.8          & 0.85         \\
                       & P-DQN     & 1           & 0.7          & 0.75         \\
                       & UniWeight & 0.8         & 0.6          & 0.65         \\
                       & SeqModel  & 0.6         & 0.6          & 0.75         \\ \midrule
\multirow{4}{*}{Bank}  & Ours      & 0.8         & 0.6          & 0.75         \\
                       & P-DQN     & 0.8         & 0.5          & 0.65         \\
                       & UniWeight & 0.6         & 0.5          & 0.55         \\
                       & SeqModel  & 0.4         & 0.4          & 0.45         \\ \bottomrule
\end{tabular}%
}
\end{table}

\begin{figure*}[htbp]
\centering
\subfloat[Higher credit behavior in Cloud]{
\label{fig:case_study11}
\includegraphics[width=0.45\linewidth]{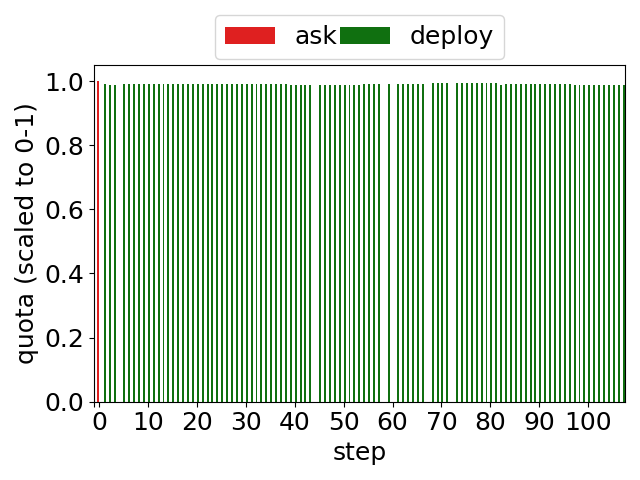}%
}%
\subfloat[Lower credit behavior in Cloud]{
\label{fig:case_study12}
\includegraphics[width=0.45\linewidth]{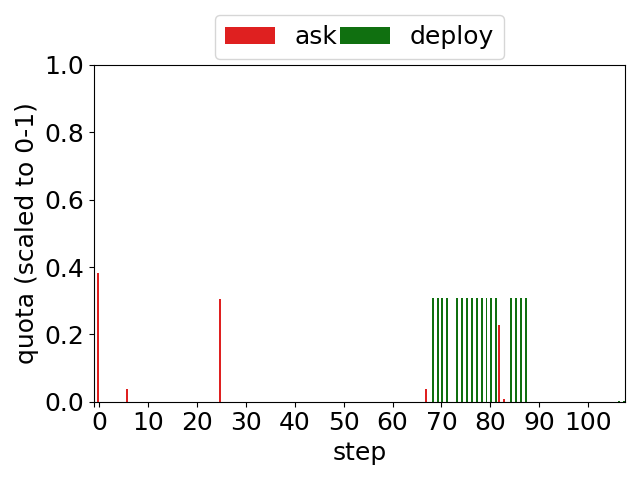}%
}\\
\subfloat[Higher credit behavior in Bank]{
\label{fig:case_study21}
\includegraphics[width=0.45\linewidth]{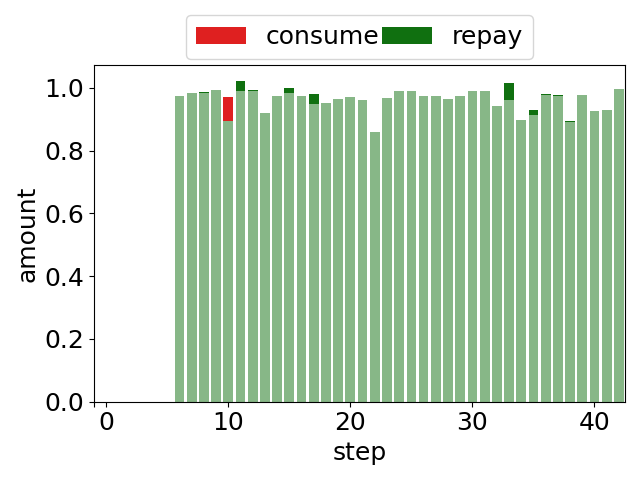}%
}%
\subfloat[Lower credit behavior in Bank]{
\label{fig:case_study22}
\includegraphics[width=0.45\linewidth]{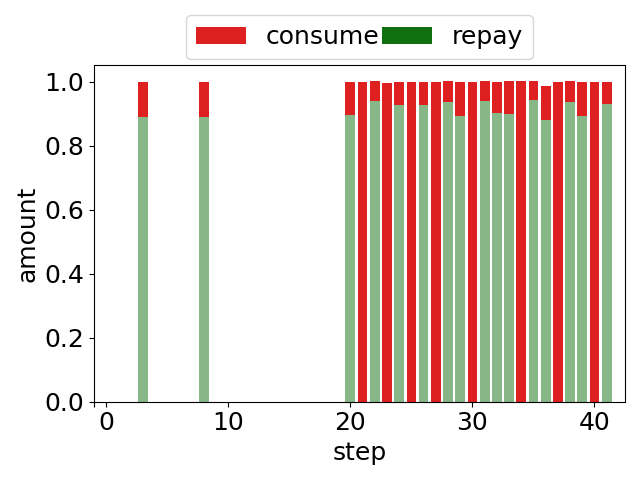}%
}
\caption{Real behaviors evaluated by our scoring system. Light green 
denote the overlap amount of actions.}
\label{fig:case_study}
\end{figure*}

\subsection{Case Study: Real Deployment of the Scoring System}
{In this case study, we first use our framework to learn two effective scoring functions for the cloud platform and simulated financial credit, respectively. Then, we use these scoring functions to score the real-world users on their real-world behaviors. These real behaviors can be accessed by two real datasets mentioned in Section \ref{subsec:aba}. We} present the real user activity records with a higher score and a lower score in Figure \ref{fig:case_study}. {In Figure \ref{fig:case_study11}, the scoring system finds a good user with a higher score and we can see that he/she just asks for the computing quota at the very beginning and then he/she deploys them as much as possible in the most timestamps, and has very less idle time. However, in Figure \ref{fig:case_study12}, a user scored very lowly by our scoring system has more invalid waiting time and asks for quota many times but uses the quota less efficiently. Similar observations can be also found in the bank application, in Figure \ref{fig:case_study21}, we use dark green to denote users' repaying amount, and red to consuming amount. The light green denotes the overlap amount of these two actions. We can see that a user ranked higher by our scoring system usually have better repayment and consumption records. He/she consumes nearly every month and repays his/her bills timely. Even sometimes he/she can not fully repay the monthly bill (like step 10), he/she still tries his/her best to repay at a later time (despite extra interest). In contrast, Figure \ref{fig:case_study22} presents a user with a very low score given by our system and we can find that he repays less and even have many non-payment records. These observations fit with our intuition very well, which can further demonstrate that our proposed method can learn a sensible scoring system automatically. }

%% file: related_work.tex
\section{related work}\label{sec:related}
{

 \textbf{Scoring Systems in Financial Areas.} User scoring systems were first used as the bank credit scoring to redefine credit-worthiness \cite{glover2018creditworthy}. The scoring function for such area was designed in a top-down manner where each factor, each parameter, and the calculation method are all determined by expert knowledge, and the function is still updated periodically \cite{zhang2010application}. With the development of machine learning techniques, there are recent works \cite{nana2022game,dumitrescu2022machine} using machine learning models on different tasks related to financial credit scoring systems like loan rejection  \cite{maldonado2010semi}, and more advanced work \cite{lei2020generative,wang2020using} to study various repayment behaviors. 

\noindent \textbf{Scoring System in Other Areas.} Beyond the financial area, recent years also witness such scoring systems applied to other areas. In online forum management,
Ganesh et al. \cite{ghalme2018design} proposed a coalition-resistant credit score function to prevent users from manipulating their credit scores by forming coalitions. In web services, user scoring is used as a part of their profiles so that the platform can better understand users for their search results\cite{pansocial2019}. There are other fields like online spammer detection and psychological analysis, which might take similar techniques to judge whether an online user is abnormal \cite{Liu_2019,Wu_2020,Dou_2020}, or one's personality traits fit their behavior patterns\cite{Sun_2022,Zhao_2018,Sun_2023}. For example, Wu et al. \cite{Wu_2020} take some observations from social spammers and then design a graph neural network to predict the possibility of spammer. Here the possibility can be treated as an abnormal score to judge the spammer. Besides, Sun et al.\cite{Sun_2022} develop a multi-modal framework to recognize users' interview videos and evaluate their psychological on five personality categories. 


}

%% file: conclusion.tex
\section{Conclusion}\label{sec:con}
\label{sec:discussion}


In this paper, we explore a self-contained framework to help design a better user scoring system. This learning problem is formulated as the adversarial learning problem between the empirical criteria attacker and an enhancer of the scoring function. Extensive experiments on two real datasets demonstrate that our proposed framework can learn robust scoring functions across different scenarios.

%% file: appendix.tex
\section{appendix}\label{sec:app}
\subsection{A Nutshell of Cloud Platform Scoring}\label{app:azure}

\subsubsection{Scoring Function}

\begin{equation}
\begin{aligned}
    s_t=&w_1 \cdot m_{\mathit{ask}}(\cdot)+w_2\cdot m_{\mathit{dep}}(\cdot)\\&+w_3 \cdot m_{\mathit{uti}}(\cdot)+w_4\cdot m_{\mathit{af}}(\cdot)+w_5\cdot m_{\mathit{df}}(\cdot)
\end{aligned}
\end{equation}

In this function, the user has 3 types of actions: \emph{ask} some quota, \emph{deploy} some quota, and \emph{wait}. Let $\mathcal{H}^t$ be the user's latest action records, which can be treated as a truncated history with a traceback window $|\mathcal{H}^t|$. For each action in $\mathcal{H}^t$, it can be denoted as a tuple $(a_i, q_i)$ where $a_i$ denotes the action type and $q_i$ is the corresponding quota. For example, $a_i=0$ means this action is "\textit{asking $q_i$ quota}"; $a_i=1$ means "\textit{deploying $q_i$ quota}", and $a_i=2$ means "\textit{waiting for nothing}" ($q_i=0$); $I(\cdot)$ is an indicator; $\lambda>0$ measures the degree of time decay. A larger $\lambda$ means the factor focuses more on recent choices. $\gamma>0$ is a scaled parameter. $q^{max}_{i}$ is max available quota when action $a_i$ happens in $\mathcal{H}^t$.

To improve the utilization, we wish the user can ask for less quota and deploy the quota as much as possible. Then we can design three factors: ask factor $m_{ask}$, deploy factor $m_{dep}$, and utilization factor $m_{uti}$ to encourage users to improve their utilization:
\begin{footnotesize} 
\begin{equation}
\begin{aligned}
&        m_{ask}\left(t, \{-\gamma_{ask},\lambda_{ask}\},\mathcal{H}^t\right)=\\
&        \tanh\left(-\gamma_{ask}\cdot \frac{\sum_{i=0}^{|\mathcal{H}^t|}q_i\cdot e^{-\lambda_{ask}\left(|\mathcal{H}^t|-i-1\right)}\cdot I\left(a_i=0\right)}{\sum_{i=0}^{|\mathcal{H}^t|} I\left(a_i=0\right)}\right)+1
\end{aligned}
\end{equation}
\end{footnotesize}

\begin{equation}
\begin{aligned}
&m_{dep}\left(t, \{\gamma_{dep},\lambda_{dep}\},\mathcal{H}^t\right)=\\
&\tanh\left(\gamma_{dep}\cdot \frac{\sum_{i=0}^{|\mathcal{H}^t|}q_i\cdot e^{-\lambda_{dep}\left(|\mathcal{H}^t|-i-1\right)}\cdot I\left(a_i=1\right)}{\sum_{i=0}^{|\mathcal{H}^t|} I\left(a_i=1\right)}\right)
\end{aligned}
\end{equation}
\begin{equation}
\begin{aligned}
& m_{uti}\left(t, \{\gamma_{uti},\lambda_{uti}\},\mathcal{H}^t\right)=\\
&\tanh\left(\gamma_{uti}\cdot \frac{\sum_{i=0}^{|\mathcal{H}^t|}q^{dep}_{i}/q^{max}_{i}\cdot e^{-\lambda_{uti}\left(|\mathcal{H}^t|-i-1\right)}}{\sum_{i=0}^{|\mathcal{H}^t|} e^{-\lambda_{uti}\left(|\mathcal{H}^t|-i-1\right)}} \right)
\end{aligned}
\end{equation}

To improve users' activeness, we wish the user can take effective actions as many as possible. We design two factors to evaluate the user's activity: ask frequency factor $m_{af}$, and deploy frequency factor $s_{df}$:
    \begin{equation}
\begin{aligned}
    &m_{af}\left(t, \{\gamma_{af},\lambda_{af}\},\mathcal{H}^t\right)=\\
    &\tanh\left(-\gamma_{af}\cdot \frac{\sum_{i=0}^{|\mathcal{H}^t|}I\left(a_i=0\right)\cdot e^{-\lambda_{af}\left(|\mathcal{H}^t|-i-1\right)}}{\sum_{i=0}^{|\mathcal{H}^t|} e^{-\lambda_{af}\left(|\mathcal{H}^t|-i-1\right)}}\right)+1
\end{aligned}
\end{equation}
\begin{equation}
\begin{aligned}
    &m_{df}\left(t, \{\gamma_{df},\lambda_{df}\},\mathcal{H}^t\right)=\\
    &\tanh\left(\gamma_{df}\cdot \frac{\sum_{i=0}^{|\mathcal{H}^t|}I\left(a_i=1\right)\cdot e^{-\lambda_{df}\left(|\mathcal{H}^t|-i-1\right)}}{\sum_{i=0}^{|\mathcal{H}^t|} e^{-\lambda_{df}\left(|\mathcal{H}^t|-i-1\right)}}\right)
\end{aligned}
\end{equation}

{
\subsubsection{Reward}
The reward of the counter-empirical attacker:
\begin{equation}
    r_t=\left\{
    \begin{aligned}
            -0.01& & \text{ if } q_t^{dep}>q_{t-1}^{dep}, s_t>s_{t-1}\\
            1.1*|\Delta s|& & \text{ if } q_t^{dep}>q_{t-1}^{dep}, s_t\leq s_{t-1}\\
            -0.02 && \text{ if } q_t^{dep}\leq q_{t-1}^{dep}, s_t<s_{t-1}\\
             1.1*|\Delta s| && \text{ if } q_t^{dep}\leq q_{t-1}^{dep}, s_t\geq s_{t-1}\\
             -0.06 && \text{ if } q_t^{ask}>0, s_t<s_{t-1}\\
             1.1*|\Delta s|&& \text{ if } q_t^{ask}>0, s_t\geq s_{t-1}\\
             0 && \text{ otherwise }\\
        \end{aligned}
        \right.
\end{equation}
where $\Delta s=s_{t}-s_{t-1}$. Note that each time only one condition occurs in the above function.  
}

\subsection{A Nutshell of Financial Credit Scoring}\label{app:bank}
{\subsubsection{Scoring Function}}
\begin{equation}
    s^t=w_1 m_{\mathit{rep}}(\cdot)+w_2 m_{\mathit{con}}(\cdot)+w_3 m_{\mathit{tra}}(\cdot)+w_4 m_{\mathit{wai}}(\cdot)
\end{equation}

Here, users have four action types: \emph{repay} ($a_i=0$), \emph{consume} ($a_i=1$), \emph{in-out} ($a_i=2$), and \emph{inactivity} ($a_i=3$).
Let $q^{\mathit{rep}}$ and $q^{con}$ denote the repaid amount and consumed amount respectively. We consider four factors to encourage good behavior: repay factor $m_{\mathit{rep}}$, consume factor $m_{con}$, trade-off factor $m_{\mathit{tra}}$, wait factor $m_{\mathit{wai}}$:
\begin{equation}
\begin{aligned}
&m_{\mathit{rep}}\left(t, \{\gamma_{\mathit{rep}},\lambda_{\mathit{rep}}\},\mathcal{H}^t\right)=\\ &\tanh\left(\gamma_{\mathit{rep}} \sum_{i=0}^{|\mathcal{H}^t|}I(a_i=0)q_i^{\mathit{rep}}\cdot e^{-\lambda_{\mathit{rep}}\left(|\mathcal{H}^t|-i-1\right)}\right)
\end{aligned}
\end{equation}

\begin{equation}
\begin{aligned}
&m_{\mathit{con}}\left(t, \{\gamma_{\mathit{con}},\lambda_{\mathit{con}}\},\mathcal{H}^t\right)=\\ &\tanh\left(-\gamma_{\mathit{con}} \sum_{i=0}^{|\mathcal{H}^t|}I(a_i=1)q_i^{\mathit{con}}\cdot e^{-\lambda_{\mathit{con}}\left(|\mathcal{H}^t|-i-1\right)}\right)+1
\end{aligned}
\end{equation}

\begin{equation}
\begin{aligned}
    & m_{\mathit{tra}}\left(t, \{\gamma_{\mathit{tra}},\lambda_{\mathit{tra}}\},\mathcal{H}^t\right)=\\
    & \tanh\!\left(\!-\gamma_{\mathit{tra}} \sum_{i=0}^{|\mathcal{H}^t|}I(a_i\!=2)|\frac{q_i^{\mathit{con}}}{q_i^{\mathit{rep}}}\!-1|\cdot e^{-\lambda_{\mathit{tra}}\left(|\mathcal{H}^t|-i-1\right)}\!\right)\!+\!1
\end{aligned}
\end{equation}
\begin{equation}
\begin{aligned}
&m_{\mathit{wai}}\left(t, \{\gamma_{\mathit{wai}},\lambda_{\mathit{wai}}\},\mathcal{H}^t\right)=\\ &\tanh\left(-\gamma_{\mathit{wai}} \sum_{i=0}^{|\mathcal{H}^t|}I(a_i=3)\cdot e^{-\lambda_{\mathit{wai}}\left(|\mathcal{H}^t|-i-1\right)}\right)+1
\end{aligned}
\end{equation}
{
\subsubsection{Reward}
\begin{equation}
    r_t=\left\{
    \begin{aligned}
            -q^{rep}* |\Delta s| & & \text{ if } a_t=0, s_t>s_{t-1}\\
            q^{rep}* |\Delta s| & & \text{ if } a_t=0, s_t\leq s_{t-1}\\
            q^{con}* |\Delta s|  & & \text{ if } a_t=1, s_t>s_{t-1}\\
            -q^{con}* |\Delta s|  &&  \text{ if } a_t=1, s_t\leq s_{t-1}\\
            |q^{con}/q^{rep}-1|* |\Delta s|  & & \text{ if } a_t=2, s_t>s_{t-1}\\
            -|q^{con}/q^{rep}-1|* |\Delta s|  & & \text{ if } a_t=2, s_t \leq s_{t-1}\\
             0 &&  \text{ otherwise }\\
        \end{aligned}
        \right.
\end{equation}
}

%% file: author.tex

\begin{IEEEbiographynophoto}{Xiangguo Sun} is now working as a postdoctoral research fellow at the Chinese University of Hong Kong under the supervision of Prof. Hong Cheng. He was recognized as the "Social Computing Rising Star" in 2023 from CAAI. He studied at  Zhejiang Lab as a visiting researcher hosted by Prof. Hongyang Chen  in 2022. In the same year, he received his Ph.D. from Southeast University under the supervision of Prof. Bo Liu and won the Distinguished Ph.D. Dissertation Award. During his Ph.D. study, he worked as a research intern at Microsoft Research Asia from Sep 2021 to Jan 2022, and won the ''Award of Excellence'' from MSRA in the Stars of Tomorrow Program. He also studied as a joint Ph.D. student at The University of Queensland hosted by ARC Future Fellow Prof. Hongzhi Yin from Sep 2019 to Sep 2021, Australia. His research interests include social computing and network learning. He was the winner of the Best Research Paper Award at KDD'23. This is also the first time a university from Mainland China, Hong Kong, and Macau has received this award in the KDD history. He served as the PC member/reviewer of NeurIPS 2023, The Web Conference (WWW) 2023, SIGIR 2021, ICDE 2021, SIGKDD (2020, 2021, 2022, 2023), VLDB 2022, DASFAA (2020, 2023), TNNLS, etc. He has published 11 CORE A*, 9 CCF A, and 13 SCI (including 6 IEEE Trans), some of which appear in SIGKDD, VLDB, The Web Conference (WWW), TKDE, TOIS, WSDM, TNNLS, CIKM, etc.
\end{IEEEbiographynophoto}	
\vspace{-10 mm}

\begin{IEEEbiographynophoto}{Hong Cheng} is a Professor in the Department of Systems Engineering and Engineering Management, Chinese University of Hong Kong. She received the Ph.D. degree from the University of Illinois at Urbana-Champaign in 2008. Her research interests include data mining, database systems, and machine learning. She received research paper awards at ICDE'07, SIGKDD'06, and SIGKDD'05, and the certificate of recognition for the 2009 SIGKDD Doctoral Dissertation 
 Award. She received the 2010 Vice-Chancellor's Exemplary Teaching Award at the Chinese University of Hong Kong. 
\end{IEEEbiographynophoto}
\vspace{-10 mm}

\begin{IEEEbiographynophoto}{Hang Dong} is a Researcher in the DKI (Data, Knowledge, Intelligence) Group of Microsoft Research Asia. She received both her B.S and Ph.D from Tsinghua University in 2015 and 2020. Her research interest include statistical network modeling, network representation learning and data-driven service intelligence especially in cloud services.
\end{IEEEbiographynophoto}
\vspace{-10 mm}
\begin{IEEEbiographynophoto}{Bo Qiao} is a Research SDE in the DKI (Data, Knowledge, Intelligence) area at Microsoft Research Asia. His research interests include software engineering and cloud services.
\end{IEEEbiographynophoto}
\vspace{-10 mm}
\begin{IEEEbiographynophoto}{Si Qin} is a Principal Researcher and Research Manager in the Data, Knowledge, Intelligence (DKI) area of Microsoft Research Asia (MSRA). He has been dedicated to AI for IT Operations (AIOps) research and conducting decision-making for building high-quality cloud services based on signal processing, machine learning, and optimization technologies. The research technologies have been successfully transferred into multiple Microsoft product divisions, such as Microsoft Azure and Office 365.
\end{IEEEbiographynophoto}
\vspace{-10 mm}
\begin{IEEEbiographynophoto}{Qingwei Lin} is a Sr. Principal Research Manager in the DKI (Data, Knowledge, Intelligence) area of Microsoft Research Asia. He is leading a team of researchers working on data-driven technologies for cloud intelligence, with innovations in machine learning and data mining algorithms. In cloud intelligence area, Qingwei has multiple publications in the conferences of AAAI, IJCAI, SigKDD, WWW, ICSE, FSE, ASE, OSDI, NSDI, USENIX ATC, etc. The research technologies have been transferred into multiple Microsoft product divisions, such as Microsoft Azure, Office365, Windows, Bing, etc. Qingwei hosted Microsoft company-wide “Cloud Service Intelligence Summit” as the Chair for 4 consecutive years. He joined Microsoft Research in 2006. 
\end{IEEEbiographynophoto}

\vfill

%% file: IEEE_Trans_main.bbl
\begin{thebibliography}{10}
\providecommand{\url}[1]{#1}
\csname url@samestyle\endcsname
\providecommand{\newblock}{\relax}
\providecommand{\bibinfo}[2]{#2}
\providecommand{\BIBentrySTDinterwordspacing}{\spaceskip=0pt\relax}
\providecommand{\BIBentryALTinterwordstretchfactor}{4}
\providecommand{\BIBentryALTinterwordspacing}{\spaceskip=\fontdimen2\font plus
\BIBentryALTinterwordstretchfactor\fontdimen3\font minus
  \fontdimen4\font\relax}
\providecommand{\BIBforeignlanguage}[2]{{%
\expandafter\ifx\csname l@#1\endcsname\relax
\typeout{** WARNING: IEEEtran.bst: No hyphenation pattern has been}%
\typeout{** loaded for the language `#1'. Using the pattern for}%
\typeout{** the default language instead.}%
\else
\language=\csname l@#1\endcsname
\fi
#2}}
\providecommand{\BIBdecl}{\relax}
\BIBdecl

\bibitem{wang2020using}
W.~Wang, C.~Lesner, A.~Ran, M.~Rukonic, J.~Xue, and E.~Shiu, ``Using small
  business banking data for explainable credit risk scoring,'' \emph{AAAI},
  vol.~34, no.~08, pp. 13\,396--13\,401, 2020.

\bibitem{teerasoponpong2022decision}
S.~Teerasoponpong and A.~Sopadang, ``Decision support system for adaptive
  sourcing and inventory management in small-and medium-sized enterprises,''
  \emph{Robotics and Computer-Integrated Manufacturing}, vol.~73, p. 102226,
  2022.

\bibitem{israel2014credit}
S.~Israel, A.~Caspi, D.~W. Belsky, H.~Harrington, S.~Hogan, R.~Houts,
  S.~Ramrakha, S.~Sanders, R.~Poulton, and T.~E. Moffitt, ``Credit scores,
  cardiovascular disease risk, and human capital,'' \emph{PNAS}, vol. 111,
  no.~48, pp. 17\,087--17\,092, 2014.

\bibitem{guo2016personal}
G.~Guo, F.~Zhu, E.~Chen, L.~Wu, Q.~Liu, Y.~Liu, and M.~Qiu, ``Personal credit
  profiling via latent user behavior dimensions on social media,'' in
  \emph{PAKDD}.\hskip 1em plus 0.5em minus 0.4em\relax Springer, 2016, pp.
  130--142.

\bibitem{maldonado2010semi}
S.~Maldonado and G.~Paredes, ``A semi-supervised approach for reject inference
  in credit scoring using svms,'' in \emph{ICDM}.\hskip 1em plus 0.5em minus
  0.4em\relax Springer, 2010, pp. 558--571.

\bibitem{li2009hybrid}
F.-C. Li, ``The hybrid credit scoring strategies based on knn classifier,'' in
  \emph{Sixth International Conference on Fuzzy Systems and Knowledge
  Discovery}, vol.~1.\hskip 1em plus 0.5em minus 0.4em\relax IEEE, 2009, pp.
  330--334.

\bibitem{thomas2017credit}
L.~Thomas, J.~Crook, and D.~Edelman, \emph{Credit scoring and its
  applications}.\hskip 1em plus 0.5em minus 0.4em\relax SIAM, 2017.

\bibitem{dumitrescu2022machine}
E.~Dumitrescu, S.~Hue, C.~Hurlin, and S.~Tokpavi, ``Machine learning for credit
  scoring: Improving logistic regression with non-linear decision-tree
  effects,'' \emph{European Journal of Operational Research}, vol. 297, no.~3,
  pp. 1178--1192, 2022.

\bibitem{kozodoi2022fairness}
N.~Kozodoi, J.~Jacob, and S.~Lessmann, ``Fairness in credit scoring:
  Assessment, implementation and profit implications,'' \emph{European Journal
  of Operational Research}, vol. 297, no.~3, pp. 1083--1094, 2022.

\bibitem{bhatnagar2007incremental}
S.~Bhatnagar, M.~Ghavamzadeh, M.~Lee, and R.~S. Sutton, ``Incremental natural
  actor-critic algorithms,'' \emph{Advances in neural information processing
  systems}, vol.~20, pp. 105--112, 2007.

\bibitem{grondman2012survey}
I.~Grondman, L.~Busoniu, G.~A. Lopes, and R.~Babuska, ``A survey of
  actor-critic reinforcement learning: Standard and natural policy gradients,''
  \emph{IEEE Transactions on Systems, Man, and Cybernetics, Part C
  (Applications and Reviews)}, vol.~42, no.~6, pp. 1291--1307, 2012.

\bibitem{sutton2018reinforcement}
R.~S. Sutton and A.~G. Barto, \emph{Reinforcement learning: An
  introduction}.\hskip 1em plus 0.5em minus 0.4em\relax MIT press, 2018.

\bibitem{schulman2015high}
J.~Schulman, P.~Moritz, S.~Levine, M.~I. Jordan, and P.~Abbeel,
  ``High-dimensional continuous control using generalized advantage
  estimation,'' in \emph{ICLR}, 2016.

\bibitem{schulman2017proximal}
J.~Schulman, F.~Wolski, P.~Dhariwal, A.~Radford, and O.~Klimov, ``Proximal
  policy optimization algorithms,'' \emph{arXiv preprint arXiv:1707.06347},
  2017.

\bibitem{cuturi2017soft}
M.~Cuturi and M.~Blondel, ``Soft-dtw: a differentiable loss function for
  time-series,'' in \emph{ICML}.\hskip 1em plus 0.5em minus 0.4em\relax PMLR,
  2017, pp. 894--903.

\bibitem{Littman1994game}
M.~L. littman, ``Markov games as a framework for multi-agent reinforcement
  learning,'' in \emph{ICML}, ser. ICML'94.\hskip 1em plus 0.5em minus
  0.4em\relax San Francisco, CA, USA: Morgan Kaufmann Publishers Inc., 1994, p.
  157–163.

\bibitem{perolat2015approximate}
J.~P{\'{e}}rolat, B.~Scherrer, B.~Piot, and O.~Pietquin, ``Approximate dynamic
  programming for two-player zero-sum markov games,'' in \emph{Proceedings of
  the 32nd International Conference on Machine Learning, {ICML} 2015, Lille,
  France, 6-11 July 2015}, ser. {JMLR} Workshop and Conference Proceedings,
  F.~R. Bach and D.~M. Blei, Eds., vol.~37.\hskip 1em plus 0.5em minus
  0.4em\relax JMLR.org, 2015, pp. 1321--1329.

\bibitem{xiong2018parametrized}
J.~Xiong, Q.~Wang, Z.~Yang, P.~Sun, L.~Han, Y.~Zheng, H.~Fu, T.~Zhang, J.~Liu,
  and H.~Liu, ``Parametrized deep q-networks learning: Reinforcement learning
  with discrete-continuous hybrid action space,'' \emph{arXiv preprint
  arXiv:1810.06394}, 2018.

\bibitem{glover2018creditworthy}
L.~Glover, ``Creditworthy: A history of consumer surveillance and financial
  identity in america by josh lauer,'' \emph{Journal of Intellectual Freedom \&
  Privacy}, vol.~2, no. 3-4, pp. 33--34, 2018.

\bibitem{zhang2010application}
Y.~Zhang, M.~A. Orgun, R.~A. Baxter, and W.~Lin, ``An application of element
  oriented analysis based credit scoring,'' in \emph{ICDM}, P.~Perner, Ed.,
  vol. 6171.\hskip 1em plus 0.5em minus 0.4em\relax Springer, 2010, pp.
  544--557.

\bibitem{nana2022game}
Z.~Nana, W.~Xiujian, and Z.~Zhongqiu, ``Game theory analysis on credit risk
  assessment in e-commerce,'' \emph{Information Processing \& Management},
  vol.~59, no.~1, p. 102763, 2022.

\bibitem{lei2020generative}
K.~Lei, Y.~Xie, S.~Zhong, J.~Dai, M.~Yang, and Y.~Shen, ``Generative
  adversarial fusion network for class imbalance credit scoring,'' \emph{Neural
  Computing and Applications}, vol.~32, no.~12, pp. 8451--8462, 2020.

\bibitem{ghalme2018design}
G.~Ghalme, S.~Gujar, A.~Kumar, S.~Jain, and Y.~Narahari, ``Design of coalition
  resistant credit score functions for online discussion forums,'' in
  \emph{AAMAS}, 2018, pp. 95--103.

\bibitem{pansocial2019}
S.~Pan and T.~Ding, ``Social media-based user embedding: A literature review,''
  in \emph{IJCAI}, 2019.

\bibitem{Liu_2019}
B.~Liu, X.~Sun, Z.~Ni, J.~Cao, J.~Luo, B.~Liu, and X.~Fu, ``Co-detection of
  crowdturfing microblogs and spammers in online social networks,'' \emph{World
  Wide Web}, vol.~23, no.~1, pp. 573--607, oct 2019.

\bibitem{Wu_2020}
Y.~Wu, D.~Lian, Y.~Xu, L.~Wu, and E.~Chen, ``Graph convolutional networks with
  markov random field reasoning for social spammer detection,''
  \emph{Proceedings of the {AAAI} Conference on Artificial Intelligence},
  vol.~34, no.~01, pp. 1054--1061, apr 2020.

\bibitem{Dou_2020}
Y.~Dou, G.~Ma, P.~S. Yu, and S.~Xie, ``Robust spammer detection by nash
  reinforcement learning,'' in \emph{Proceedings of the 26th {ACM} {SIGKDD}
  International Conference on Knowledge Discovery {\&} Data Mining}.\hskip 1em
  plus 0.5em minus 0.4em\relax {ACM}, aug 2020.

\bibitem{Sun_2022}
X.~Sun, B.~Liu, L.~Ai, D.~Liu, Q.~Meng, and J.~Cao, ``In your eyes: Modality
  disentangling for personality analysis in short video,'' \emph{{IEEE}
  Transactions on Computational Social Systems}, pp. 1--12, 2022.

\bibitem{Zhao_2018}
S.~Zhao, G.~Ding, J.~Han, and Y.~Gao, ``Personality-aware personalized emotion
  recognition from physiological signals,'' in \emph{Proceedings of the
  Twenty-Seventh International Joint Conference on Artificial
  Intelligence}.\hskip 1em plus 0.5em minus 0.4em\relax International Joint
  Conferences on Artificial Intelligence Organization, jul 2018.

\bibitem{Sun_2023}
X.~Sun, H.~Cheng, B.~Liu, J.~Li, H.~Chen, G.~Xu, and H.~Yin, ``Self-supervised
  hypergraph representation learning for sociological analysis,'' \emph{{IEEE}
  Transactions on Knowledge and Data Engineering}, pp. 1--12, 2023.

\end{thebibliography}
